\documentclass[submission,copyright,creativecommons]{eptcs}
\usepackage{geometry}             
\pdfoutput=1

\usepackage{graphicx}
\usepackage{graphbox}
\usepackage{hyperref}
\usepackage{amssymb}
\usepackage{epstopdf}
\usepackage{algorithm}
\usepackage{amsmath}
\usepackage[noend]{algpseudocode}
\usepackage{enumitem}
\usepackage{multirow}
\usepackage{hhline}
\usepackage{tabularx}
\usepackage{longtable}
\usepackage{booktabs}
\usepackage{array}
\usepackage{pdflscape}
\usepackage[utf8]{inputenc}
\usepackage[hang,footnotesize,bf]{caption}
\usepackage[acronym]{glossaries} 

\usepackage{cleveref}
\newcommand{\mycomment}[1]{}           

\usepackage{listings}
\usepackage{color}

\definecolor{dkgreen}{rgb}{0,0.6,0}
\definecolor{gray}{rgb}{0.5,0.5,0.5}
\definecolor{mauve}{rgb}{0.58,0,0.82}
\definecolor{backcolour}{rgb}{0.95,0.95,0.92}

\lstdefinestyle{mystyle}{
    backgroundcolor=\color{backcolour},
    commentstyle=\color{codegreen},
    keywordstyle=\color{magenta},
    numberstyle=\tiny\color{codegray},
    stringstyle=\color{codepurple},
    basicstyle=\ttfamily\footnotesize,
    breakatwhitespace=false,
    breaklines=true,
    captionpos=b,
    keepspaces=true,
    numbers=left,
    numbersep=5pt,
    showspaces=false,
    showstringspaces=false,
    showtabs=false,
    tabsize=2
}
\usepackage{tikz}
\usetikzlibrary{trees, arrows.meta, positioning, mindmap, backgrounds, shadows}
\usepackage{forest}

\newacronym{ai}{AI}{artificial intelligence}
\newacronym{albert}{ALBERT}{a lite BERT}
\newacronym{api}{API}{application program interface}
\newacronym{apr}{APR}{annual percentage rate}
\newacronym{agi}{AGI}{artificial general intelligence}
\newacronym{bert}{BERT}{bidirectional encoder representations from transformers}
\newacronym{bls}{BLS}{Boneh–Lynn–Shacham}
\newacronym{bn}{BN}{Bayesian network}
\newacronym{cbeth}{cbETH}{Coinbase wrapped staked ETH}
\newacronym{cdf}{CDF}{cumulative distribution function}
\newacronym{cl}{CL}{consensus layer}
\newacronym{clm}{CLM}{causal language modelling}
\newacronym{cpt}{CPT}{conditional probability table}
\newacronym{dao}{DAO}{decentralised autonomous organisation}
\newacronym{dms}{DMS}{data management system}
\newacronym{dos}{DoS}{denial of service}
\newacronym{dvt}{DVT}{distributed validator technology}
\newacronym{eb}{EB}{effective balance}
\newacronym{ef}{EF}{Ethereum Foundation}
\newacronym{eip}{EIP}{Ethereum Improvement Proposal}
\newacronym{el}{EL}{execution layer}
\newacronym{epbs}{ePBS}{enshrined PBS}
\newacronym{ewc}{EWC}{elastic weight consolidation}
\newacronym{ffg}{FFG}{friendly finality gadget}
\newacronym{fxs}{FXS}{Frax share}
\newacronym{ghost}{GHOST}{Greedy Heaviest-Observed Sub-Tree}
\newacronym{gpt}{GPT}{Generative Pre-trained Transformer}
\newacronym{icl}{ICL}{in-context learning}
\newacronym{iot}{IoT}{Internet of Things}
\newacronym{kr}{KR}{Knowledge Representation}
\newacronym{ldo}{LDO}{Lido DAO}
\newacronym{llama}{LLaMA}{Large Language Model Meta AI}
\newacronym{llm}{LLM}{Large Language Model}
\newacronym{lmd}{LMD}{Latest message driven}
\newacronym{lora}{LoRA}{low-rank adoption}
\newacronym{mev}{MEV}{maximal extractable value}
\newacronym{mlm}{MLM}{masked language modelling}
\newacronym{lmm}{LMM}{large multimodal model}
\newacronym{mm}{MM}{MetaMask}
\newacronym{nlp}{NLP}{natural language processing}
\newacronym{nft}{NFT}{non-fungible token}
\newacronym{npt}{NPT}{node probability table}
\newacronym{oecd}{OECD}{Organisation for Economic Co-operation and Development}
\newacronym{opt}{OPT}{open pretrained transformer}
\newacronym{p2p}{PaLM}{Pathways language model}
\newacronym{palm}{p2p}{peer-to-peer}
\newacronym{pbs}{PBS}{proposer builder separation}
\newacronym{pc}{PC}{personal computer}
\newacronym{pdf}{PDF}{probability density function}
\newacronym{peft}{PEFT}{parameter efficient fine-tuning}
\newacronym{pos}{PoS}{proof of stake}
\newacronym{pr}{PR}{pull request}
\newacronym{rag}{RAG}{retrieval augmented generation}
\newacronym{roberta}{RoBERTa}{Robustly optimised BERT approach}
\newacronym{rlhf}{RLHF}{reinforcement learning with human feedback}
\newacronym{rnn}{RNN}{recurrent neural network}
\newacronym{rpl}{RPL}{Rocket Pool}
\newacronym{rtd} {RTD}{replaced token detection}
\newacronym{sft}{SFT}{supervised fine-tuning}
\newacronym{ssf}{SSF}{single-slot finality}
\newacronym{ssl}{SSL}{self-supervised learning}
\newacronym{steth}{stETH}{Lido staked Ether}
\newacronym{ups}{UPS}{uninterruptible power supply}
\newacronym{vae}{VAE}{variational encoder}
\newacronym{vrf}{VRF}{verifiable random function}
\newacronym{web3xai}{Web3xAI}{the intersection of Web3 and AI technologies}
\newacronym{xai}{XAI}{explainable AI}

\title{A Primer on Large Language Models and their Limitations}
\vspace{16pt}
\author{Sandra Johnson\thanks{Corresponding author}%
	\institute{School of Mathematical Sciences, \\ Queensland University of Technology, Australia}
	\email{sandra.johnson@qut.edu.au}
\and David Hyland-Wood
	\institute{School of Electrical Engineering and Computer Science, \\ The University of Queensland, Australia}
	\email{david@hyland-wood.org}
}

\date{\today}                                          
\begin{document}
\def\titlerunning{LLM Primer}
\def\authorrunning{S. Johnson \& D. Hyland-Wood}
\maketitle

\begin{abstract}
This  paper provides a primer on \glspl*{llm} and identifies their strengths, limitations, applications and research directions. It is intended to be useful to those in academia and industry who are interested in gaining an understanding of the key \gls*{llm} concepts and technologies, and in utilising this knowledge in both day to day tasks  and in more complex scenarios where this technology can enhance current practices and processes.
\end{abstract}

\section{Introduction}
\label{sec:introduction}
The world of \gls*{ai} is increasingly penetrating all aspects of our personal and professional lives. This proliferation of \gls*{ai} tools and applications are being met with a mixture of excitement, scepticism and even dread \cite{zirar2023}. Excitement at the seemingly endless potential of \gls*{ai} applications such as \glspl*{llm}, especially when they are integrated ``within broader systems''  \cite{eloundou_gpts_2024}, scepticism as the realisation dawns that \glspl*{llm} are in fact fallible as evidenced by hallucinations and hence not the golden bullet that can solve all problems \cite{hannigan2024,hicks_chatgpt_2024}, and a feeling of dread for those who believe that \glspl*{llm} and \gls*{ai} have the potential to detrimentally impact our lives and make people redundant \cite{zirar2023}.

The ability of some \glspl*{llm} to pass Theory of Mind (ToM) \cite{strachan_testing_2024}\cite{kosinski_evaluating_2024} and Turing Tests \cite{biever_chatgpt_2023}\cite{mei_turing_2024} suggests support for the Computational Theory of Mind (CTM), that cognition may be substrate independent. These findings challenge biological essentialism and open new avenues for creating sophisticated \gls*{ai} systems capable of human-like reasoning and interaction. Viewed another way, these studies could be taken to provide evidence for those critical of both the Turing Test and Theory of Mind tests in assessing cognition in humans and animals.

However, it should be noted that \glspl*{llm} by themselves have no self monitoring (also called phenomenal consciousness or subjective experience) or internal, updatable model of their external environment (that is, a model of itself as a being in a world). Both of these conditions are required in some reasonable theories consciousness \cite{kuhn_landscape_2024}. Those omissions alone may be taken as evidence against \glspl*{llm} having any form of consciousness as that term is currently understood.

We begin this paper by providing an overview covering the basics of \gls*{llm} in Section \ref{overviewofllms}: \nameref{overviewofllms}, in order to provide a common understanding and vocabulary for these natural language modelling approaches. 

In Section~\ref{orchestration}:~\nameref{orchestration}, we note that \glspl*{llm} may be combined with other, more traditional, information retrieval technologies to create full-featured systems that can adapt \glspl*{llm} to their own use cases.

 In Section~\ref{risksandmitigations}:~\nameref{risksandmitigations} we identify a series of common risks when working with \glspl*{llm}.
 
 \glspl*{llm}, like any technology, are tools to be used with awareness and even caution. They may produce content that we wish they would not. Language models pre-trained on large text corpora that are highly likely to contain toxic and inappropriate content, are known to pass these biases on, or worse amplify them, when generating query responses and text \cite{gallegos2024bias}. The bias can present itself in various forms such as discrimination based on race, gender, disability, nationality or religion \cite{nangia-etal-2020-crows, touvron2023llamaopenefficientfoundation}. To this end researchers developed a challenge dataset,  CrowS-Pairs, crowd sourced using Amazon Mechanical Turk (MTurk), to measure the extent of bias in \gls*{mlm} \cite{nangia-etal-2020-crows}. More recent approaches to address such biases were collected by Gallegos et al in \cite{gallegos2024bias}.

Similar to the problems of socially inappropriate content are problems related to the accessibility of legally inappropriate content in training data. Most foundational LLMs are pre-trained to avoid assisting criminal activities such as composing blackmail letters or providing instructions on how to commit crimes, but prompt engineering may be used to work around these built-in checks and balances. We address these issues and various approaches to their resolution in Section~\ref{sec:jailbreak}:~\nameref{sec:jailbreak}.
 
 Finally, we conclude by summarising the key points related to \glspl*{llm} and their usage.

\section{An Overview of LLMs}
\label{overviewofllms}

The emergence of \glspl*{llm} is preceded by an extensive body of research in language modelling \cite{wu_bloomberggpt_2023} where word sequences were scored with the aid of a probabilistic model possibly dating back to 1976 \cite{jelinek_continuous_1976, wu_bloomberggpt_2023}. \Gls*{nlp} modelling further evolved over time with  \glspl*{rnn} becoming the model of choice for autoregressive language modelling tasks such as translation. However, \glspl*{rnn} typically process one token at a time because they are designed for sequential processing, making parallel processing hard to achieve and limiting the capture of long range sequences. The latter is often referred to as the ``vanishing gradient problem'' \cite{Hochreiter1998,roodschild_new_2020}.

\glspl*{rnn} are further constrained due to their heavy use of computing resources. Although more computing power is available these days, relieving some of the constraints, the key change maker was the publication of the seminal paper ``Attention Is All You Need'' \cite{vaswani_attention_2023}. The authors introduced a model called the \textit{Transformer}, signalling the arrival of what is commonly known as `transformer architecture', which revolutionised \gls*{nlp}. As the title implies the researchers discovered that it was possible to rely solely on self-attention and feedforward layers without recurrent connections as in \glspl*{rnn}  \cite{vaswani_attention_2023}. Moreover, with the introduction of \textit{Reformer}, a revised version of Transformer, the same performance was obtained in a far more memory-efficient way and on much longer sequences of words \cite{kitaev_reformer_2020}.

\begin{figure}[htbp]
\begin{center}
\includegraphics[width=0.9\linewidth]{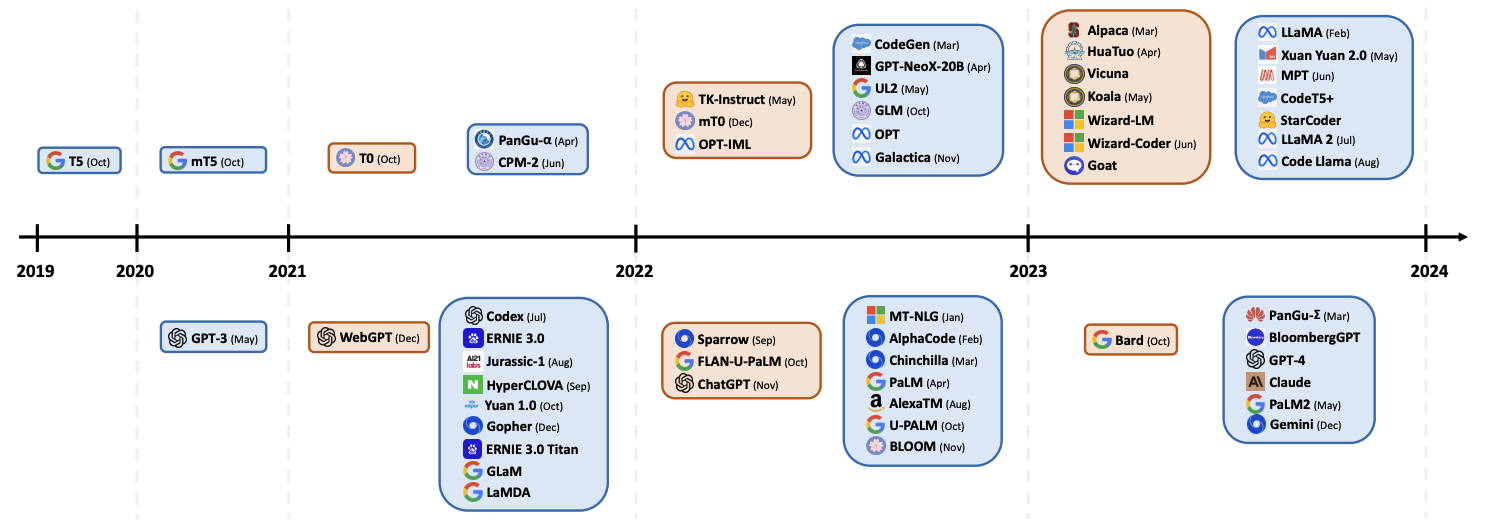}
\caption{Timeline of LLM releases: blue rounded rectangles are `pre-trained' models, while orange rectangles are `instruction-tuned' models. Models above the line indicate open-source availability, and those below the line are closed-source (image from Naveed, H., et al. \cite{naveed2024})}
\label{fig:llmreleases}
\end{center}
\end{figure}

Several \glspl*{llm} are based on this architecture, such as \gls*{bert} \cite{devlin2019bertpretrainingdeepbidirectional} and variations of \gls*{bert}: \gls*{albert}, \gls*{roberta} developed by researchers at Google; \href{https://llama.meta.com/}{\gls*{llama}} developed by Meta \cite{touvron2023llamaopenefficientfoundation} and the \gls*{gpt} models from \href{https://openai.com/}{OpenAI} such as GPT-3, GPT-4, GPT-4o, GPT-1o preview and GPT-1o mini. The proliferation of large models over time are captured in Figure \ref{fig:llmreleases} (page \pageref{fig:llmreleases}), for the period 2019 to early 2024 \cite{naveed2024}. Moreover, the diagram distinguishes between open- and closed-source models with the former above the timeline and the latter below the timeline, showing a clear trend towards open-source models \cite{naveed2024}.

Despite the increasing number of open-source models compared to closed-source models, real concerns exist for the integration of \glspl*{llm} into predominantly open source systems. Bloomberg notes \cite{bloomberg_dissecting_2024}, ``To deliver viable alternatives that compete with centralized, closed-source solutions, decentralized \gls*{ai} teams will need to innovate on model architectures and leverage model coordination platforms. In practice, this will enable ML researchers and engineers to broadly experiment with a wide variety of models aimed at different application verticals. This largely reflects how crypto networks can accelerate AI development.''

\subsection{AI Project Development utilising LLMs}
\label{sec:lifecycle}
The hype around \gls*{ai} and particularly \glspl*{llm} sparked the realisation that there are real benefits to be had if this technology can be integrated in every day personal activities and in businesses and organisations to improve productivity through automation of repetitive trivial tasks, and consequently enable employees, students and researchers to dedicate more time on interesting and complex tasks.

When embarking on an \gls*{ai}/\gls*{llm} project to harness the power of this technology, we observe that the life cycle of such a project consists mainly of four distinct phases \cite{barth_generative_nodate}:
\begin{enumerate}[noitemsep]
\item Project scoping
\item Model selection
\item Model adaption and alignment
\item Application integration
\end{enumerate}

Defining a detailed and accurate description of the use case is the first and crucial step in the project life cycle, ensuring a clear and concise scope of the project at hand.  

\begin{figure}[htbp]
\begin{center}
\includegraphics[width=0.42\linewidth]{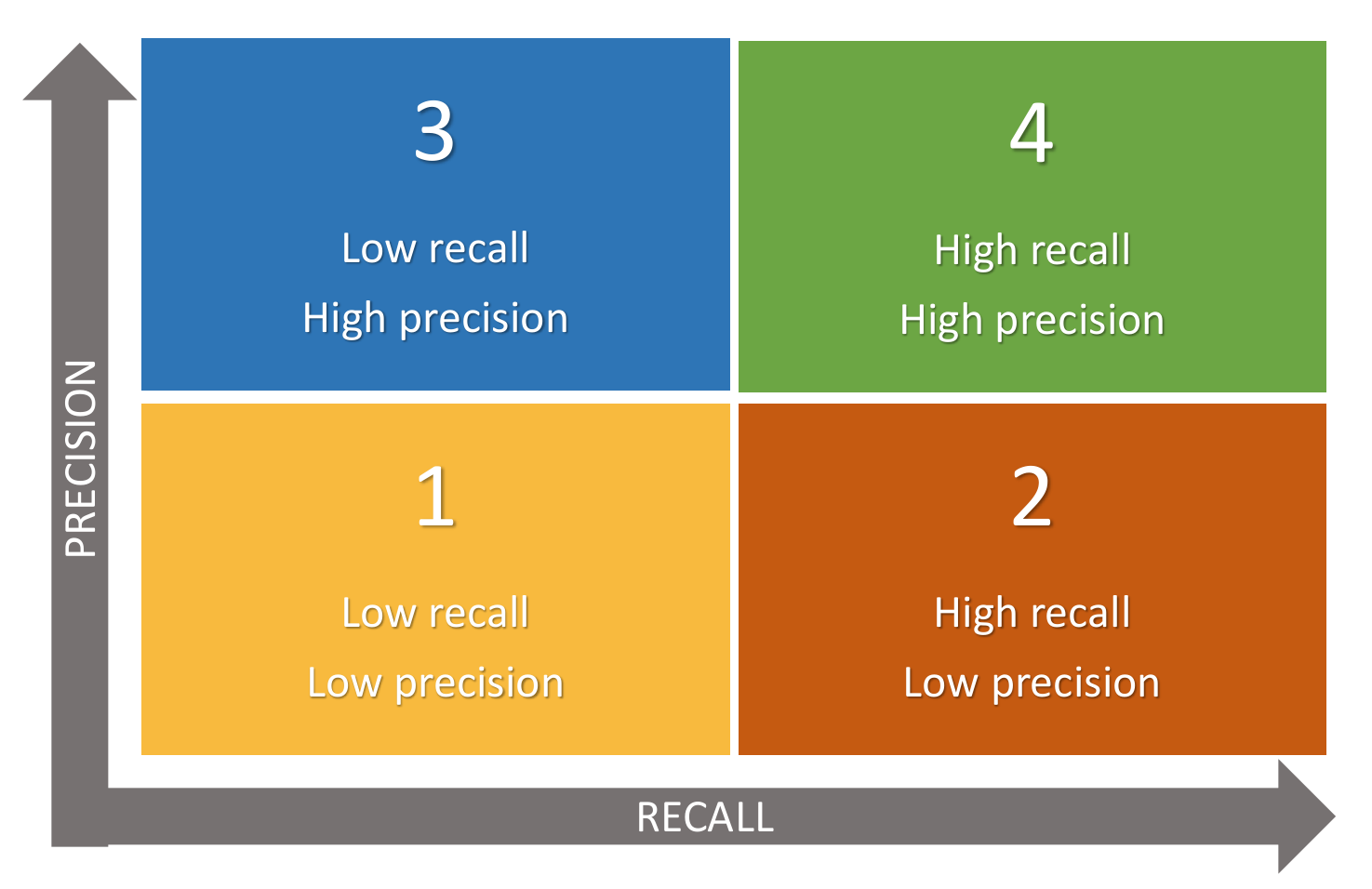}
\caption{Generative AI use case categorisation}
\label{fig:recall}
\end{center}
\end{figure}

When defining the use cases, we find that they require different degrees of precision and recall to satisfy the main task or objective. Project objectives and use cases may be viewed as roughly falling in one of the quadrants in Figure~\ref{fig:recall}, page~\pageref{fig:recall}, indicating the combination of precision and recall required for the project tasks. High precision is crucial when the output needs to be highly accurate and errors may have major consequences. Whereas high recall is essential when capturing as much relevant information as possible is the primary focus. Most use cases are likely to require at least moderate recall and precision. 
\begin{itemize}
\item \textit{Quadrant 1: Low recall, low precision}\\
Use cases in this category are typically of a creative nature, where the goal is not to get high quality or exhaustive content but simply to spark ideas and inspire. In a creative environment, generated completions may even be incomplete, vague, or somewhat nonsensical, but they can still serve as useful prompts for creative thinking. Researchers recently explored the concept of  enhancing secondary school students' creative writing skills by leveraging \gls*{ai} in the language classroom \cite{Woo2024}.  In another study researchers experimented with \gls*{ai} to aid song composition, including  the song's structure, harmony, lyrics, and hook melody \cite{Micchi2021}.

\item \textit{Quadrant 2: High recall, low precision}\\
Use cases in this quadrant are similar to the way we use search engines, e.g. searching for product support or services. Instead products such as MetaMask or Linea could integrate \gls*{ai} into customer support. Here high recall is necessary to ensure that as many customer issues as possible are addressed, while precision needs to be only moderate to low, since some responses could be reasonably generic, such as suggesting appropriate next steps for investigation or resolution. The extent to which generic advice can be given would depend on the product and the support requests \cite{Smith2024Customer}. 

\item \textit{Quadrant 3: Low recall, high precision}\\
Language translation falls into this quadrant, in the moderate to high precision and moderate recall area. In some situations it is important that the translation is accurate and precise rather than exhaustive in capturing all nuances of the source text. On the other hand, although precision is important for fluency and accuracy in certain instances, some variations in wording may be acceptable \cite{Mohan2023}.

\item \textit{Quadrant 4: High recall, high precision}\\
For use cases in healthcare and medical diagnosis it is critical to have both high precision and high recall \cite{Chu2023}. Similarly, legal tasks require high precision and moderate to high recall, depending on the task, e.g. in legal reasoning, ``generating arguments for and against particular outcomes''  would need accurate references to relevant cases and judgements, but the formulation of arguments can accomodate some creativity \cite{Ashley2017}. Nonetheless in both instances the results need be reviewed by relevant experts, before implementing any generated advice or diagnosis. These types of use cases would ultimately benefit from more complex solutions that combine \glspl*{llm} with \gls*{rag} via orchestration. We can think of use cases in this quadrant as more analogous to the way we previously heavily relied on relational databases to achieve the same ends. 

\end{itemize}

\subsection{Choosing a Foundational LLM}
\label{sec:llmchoice}

\begin{figure}[htbp]
\begin{center}
\includegraphics[width=0.6\linewidth]{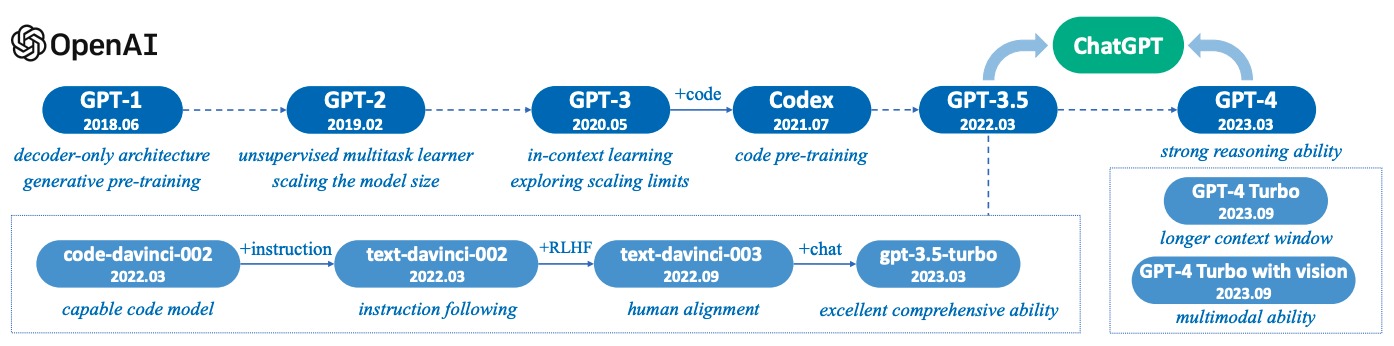}
\caption{Technical evolution of the OpenAI GPT-series models \cite{zhao_survey_2023}}
\label{fig:chatgpt}
\end{center}
\end{figure}

The choice of which \gls*{llm} to use in an AI project for a given task or functionality may appear daunting due to the sheer number of \glspl*{llm} currently available, and the emergence of a seemingly endless stream of new \glspl*{llm}, including updated versions of existing \glspl*{llm}, each purporting to improve on the previous version (Figure \ref{fig:chatgpt}, page \pageref{fig:chatgpt})  \cite{zhao_survey_2023}. Moreover,  pre-training and fine tuning techniques for models differ depending on the desired capabilities of the language model  \cite{wang_what_2022}. 

\begin{figure}[htbp]
\begin{center}
\includegraphics[width=0.6\linewidth]{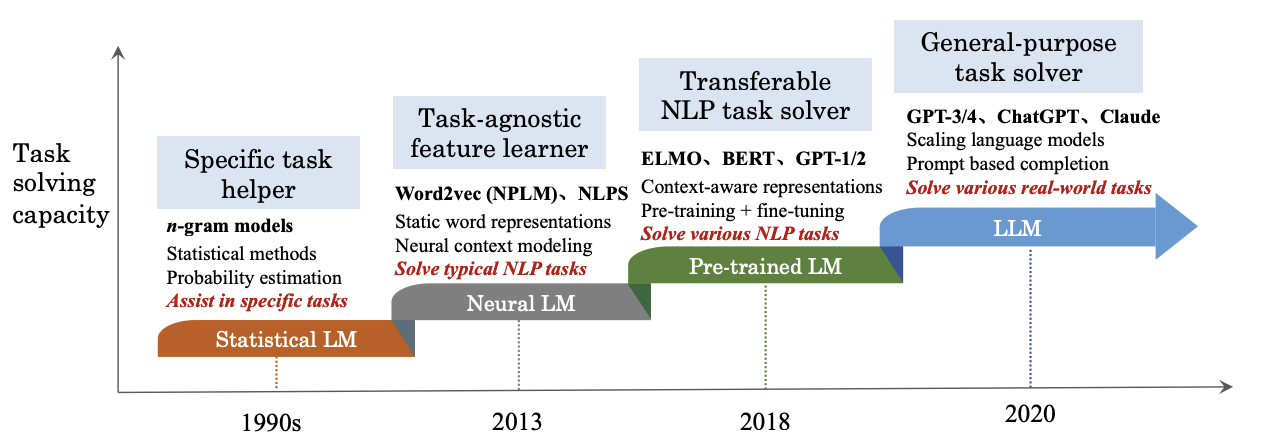}
\caption{Task solving \cite{zhao_survey_2023}}
\label{fig:tasksolving}
\end{center}
\end{figure}

\glspl*{llm} have evolved from relatively simple language tasks such as text generation to models capable of performing more complex tasks, such as those illustrated in Figure \ref{fig:tasksolving}, page \pageref{fig:tasksolving} \cite{zhao_survey_2023}.

When weighing up the options, we need to take several aspects  into consideration. The number of parameters in models vary widely and  can limit the range of devices it can run on, as well as the task objectives and applications that are best suited to a particular \gls*{llm}.  Smaller models do not necessarily perform worse than large models, especially if the model is being optimised to perform a specific task well, rather than aiming to cater for multiple use cases. Determining model size is just one consideration when choosing an existing \gls*{llm}, adapting an existing model, or building a new fit-for-purpose model from scratch. 

\begin{figure}[htbp]
\begin{center}
\includegraphics[width=0.25\linewidth]{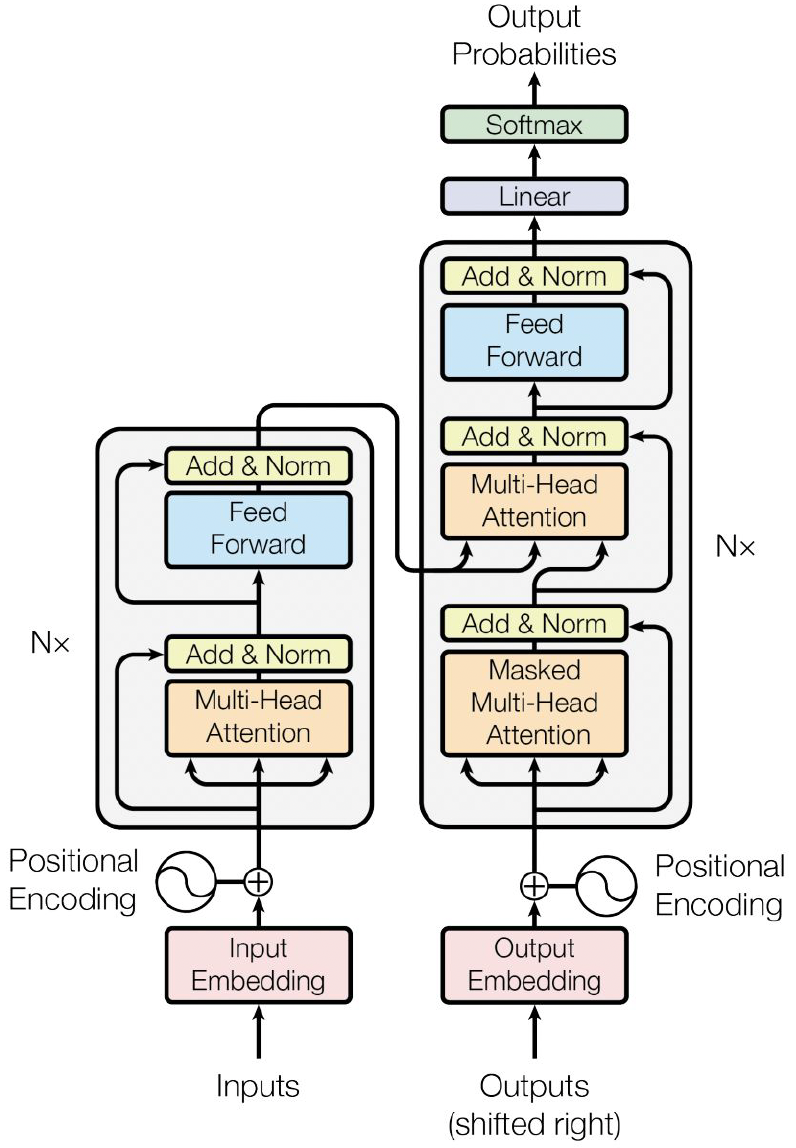} \hspace{1cm}
\includegraphics[width=0.15\linewidth]{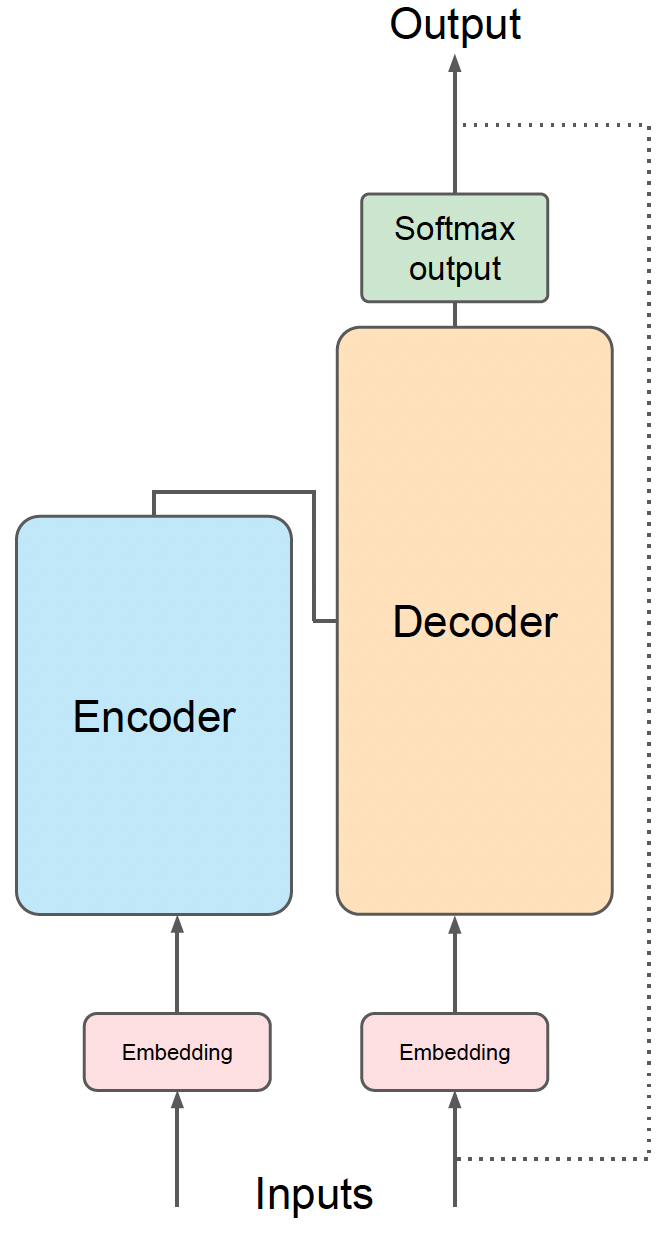} \\
(a) \hspace{5cm} (b)
\caption{(a) Attention framework (image by Vaswani, A., et al. \cite{vaswani_attention_2023}) (b)  Encoder Decoder components of transformer architecture (image by DeepLearning.AI \cite{barth_generative_nodate})}
\label{fig:transformer}
\end{center}
\end{figure}

It is helpful to keep the Transformer \gls*{llm} framework Figure~\ref{fig:transformer}(a) and high level architecture Figure~\ref{fig:transformer}(b), page \pageref{fig:transformer}, in mind when choosing a suitable foundational \gls*{llm} model. 

The encoder component encodes the model input, entered as a ``prompt'', with contextual understanding and produces one vector per input token. The decoder processes the input tokens and uses the contextual understanding from the encoder to produce new tokens \footnote{From DeepLearning.AI `Generative AI and LLMs' course}. Prompt engineering is described in Section~\ref{sec:prompt} on page~\pageref{sec:prompt}.

We can categorise foundational Transformer \gls*{llm} models as essentially one of three types: decoder-only, encoder-only, or encoder decoder models.

\subsubsection{Decoder-only models}
Decoder-only models are autoregressive models pre-trained to predict the next token based on previous tokens, making them well suited to text generation (e.g. for creative writing or content generation), autocompletion (e.g. autocompletion of sentences or lines of code), language translation, and text summarisation. 

There are several well known decoder-only \glspl*{llm}, such as the \gls*{gpt} series of models from OpenAI, \gls*{llama} and \gls*{opt} from Meta, Claude from Anthropic, \gls*{palm} from Google and Gopher from DeepMind. 

\begin{figure}[htbp]
\begin{center}
\includegraphics[width=0.6 \linewidth]{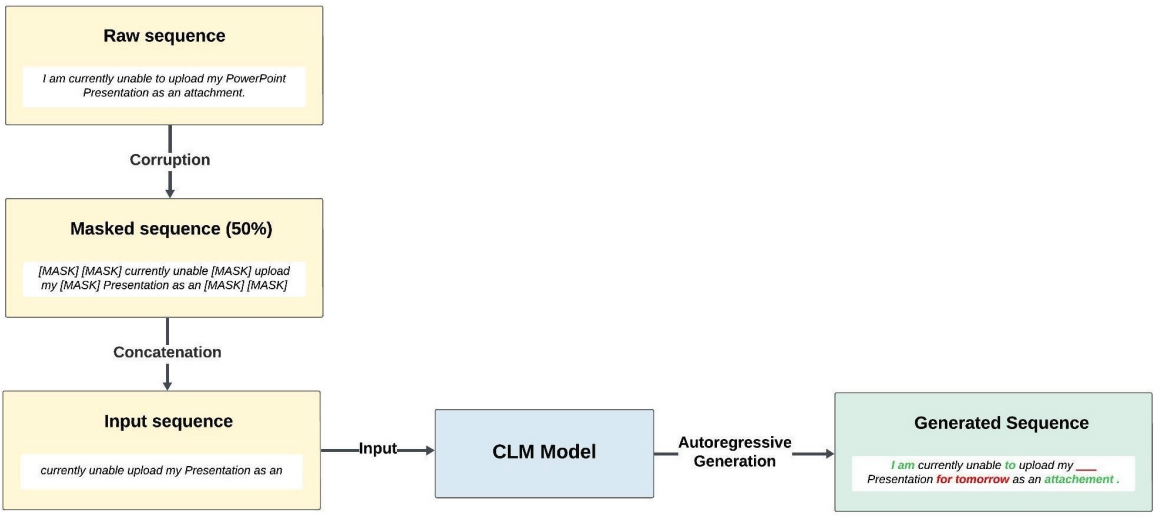} 
\caption{Causal language modelling (CLM) for decoder-only models (image by Clark, K.,et al. \cite{micheletti2024})}
\label{fig:decoder}
\end{center}
\end{figure}

\Gls*{clm}, a self-supervised learning approach, is the preferred method for training decoder-only LLMs. \Glspl*{clm} apply autoregressive modelling to input data to predict future tokens based on past tokens, an approach that is common in time series prediction and recurrent neural networks. Decoder \glspl*{llm} leverage their uni-directional, autoregressive nature to learn language patterns (Figure~\ref{fig:decoder}, page~\pageref{fig:decoder}). \Gls*{clm} token prediction is uni-directional because only the past tokens are used to predict the next tokens.

\subsubsection{Encoder-only models}
Encoder-only models are auto-encoding models, well suited to tasks that involve understanding and extracting meaning from text, such as word classification, named entity recognition, question answering and sentiment analysis. 

Key foundational encoder-only models include BERT \cite{devlin2019bertpretrainingdeepbidirectional}, variant RoBERTa, and \href{https://iclr.cc/virtual_2020/poster_r1xMH1BtvB.html}{ELECTRA} \cite{clark2020electra}. 

\begin{figure}[htbp]
\begin{center}
\includegraphics[width=0.6 \linewidth]{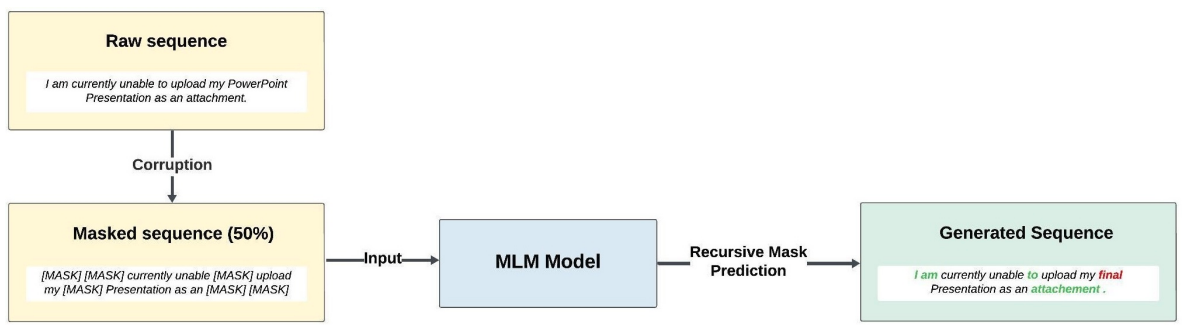} 
\caption{Masked language modelling (MLM) for encoder-only models (image by Clark, K.,et al. \cite{micheletti2024})}
\label{fig:encoder}
\end{center}
\end{figure}

\gls*{mlm}, most frequently used to  pre-train encoder models \cite{devlin2019bertpretrainingdeepbidirectional, Mohan2023}, is  a self-supervised learning technique that randomly masks tokens in an input sequence with the aim of learning the masked tokens based on the surrounding context provided by the unmasked tokens \cite{nangia-etal-2020-crows, micheletti2024}. Hence \gls*{mlm} differs from \gls*{clm} by using unmasked tokens both before and after masked tokens, providing a bi-directional understanding of context instead of being limited to the words that precede it (Figure~\ref{fig:encoder}, page~\pageref{fig:encoder}). 

Alternatively, techniques such as \gls*{rtd}, which was used to train ELECTRA, may be used \cite{clark2020electra}. With \gls*{rtd} instead of masking tokens as in \gls*{mlm}, a small generator model replaces some tokens with plausible alternatives and the encoder (discriminator) is then trained to detect the tokens that have been replaced \cite{micheletti2024}.

\subsubsection{Encoder Decoder models}
Encoder-decoder models are sequence-to-sequence models, and suited to tasks that require both understanding and generation of text, such as translation, summarisation, question answering and dialogue systems.

Two notable sequence-to-sequence \glspl*{llm} are T5 \cite{raffel_exploring_2023} and BART \cite{lewis2019bartdenoisingsequencetosequencepretraining} with both aiming to denoise corrupted inputs, via slightly different pre-training approaches.

\begin{figure}[htbp]
\begin{center}
\includegraphics[width=0.5 \linewidth]{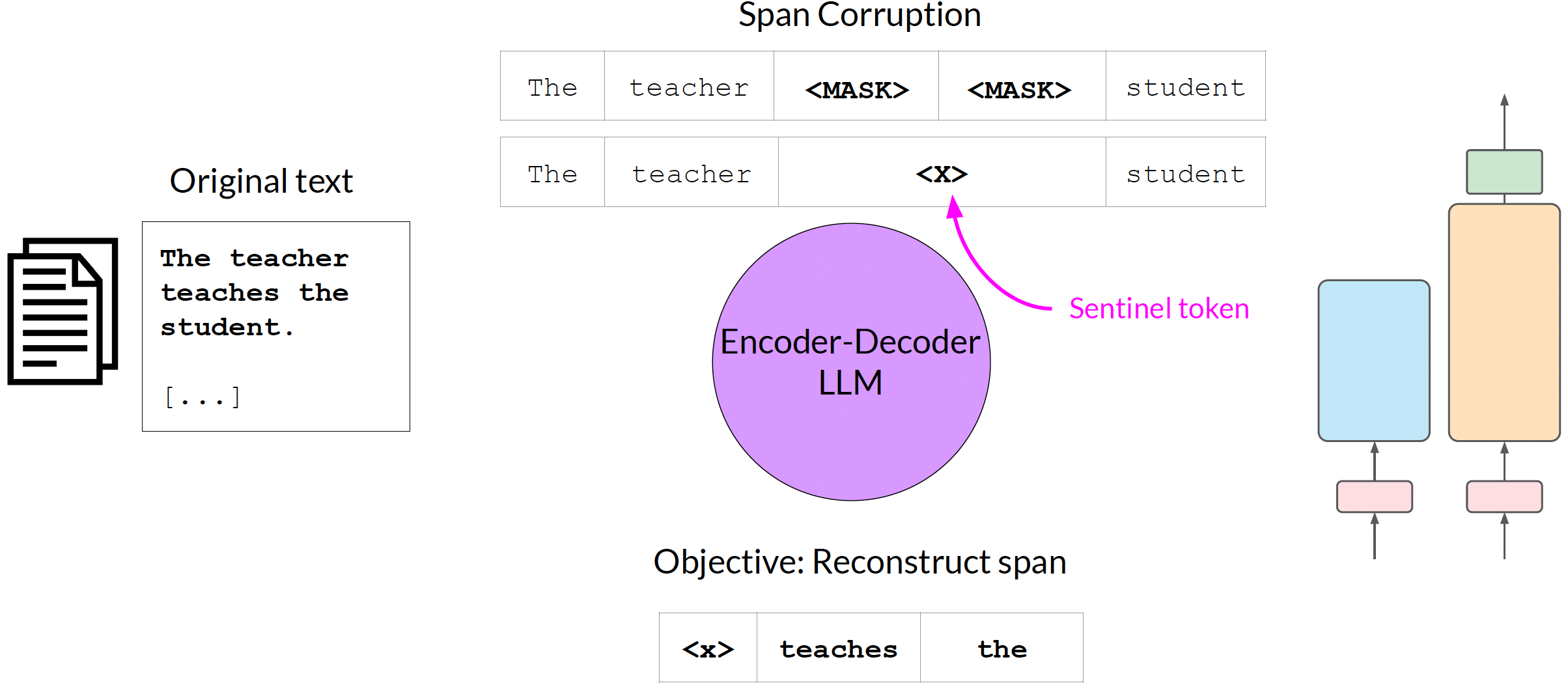} 
\caption{Encoder-decoder model (image by DeepLearning.AI \cite{barth_generative_nodate})}
\label{fig:seq2seq}
\end{center}
\end{figure}

For T5 the encoder is pre-trained using span corruption, where random sequences of tokens are masked and replaced with unique Sentinel tokens ($<x>$) that are added to the vocabulary. The decoder then reconstructs the masked token sequences in an autoregressive manner (Figure~\ref{fig:seq2seq}, page~\pageref{fig:seq2seq}). On the other hand BART uses more varied forms of corruption, including sentence permutation. Another interesting encoder-decoder model is the multilingual variant of T5, mT5 \cite{xue2021mt5}. 

\subsection{Pre-training Foundational LLMs}
\label{sec:pretraining}
\begin{figure}[htbp]
\begin{center}
\includegraphics[width=0.7 \linewidth]{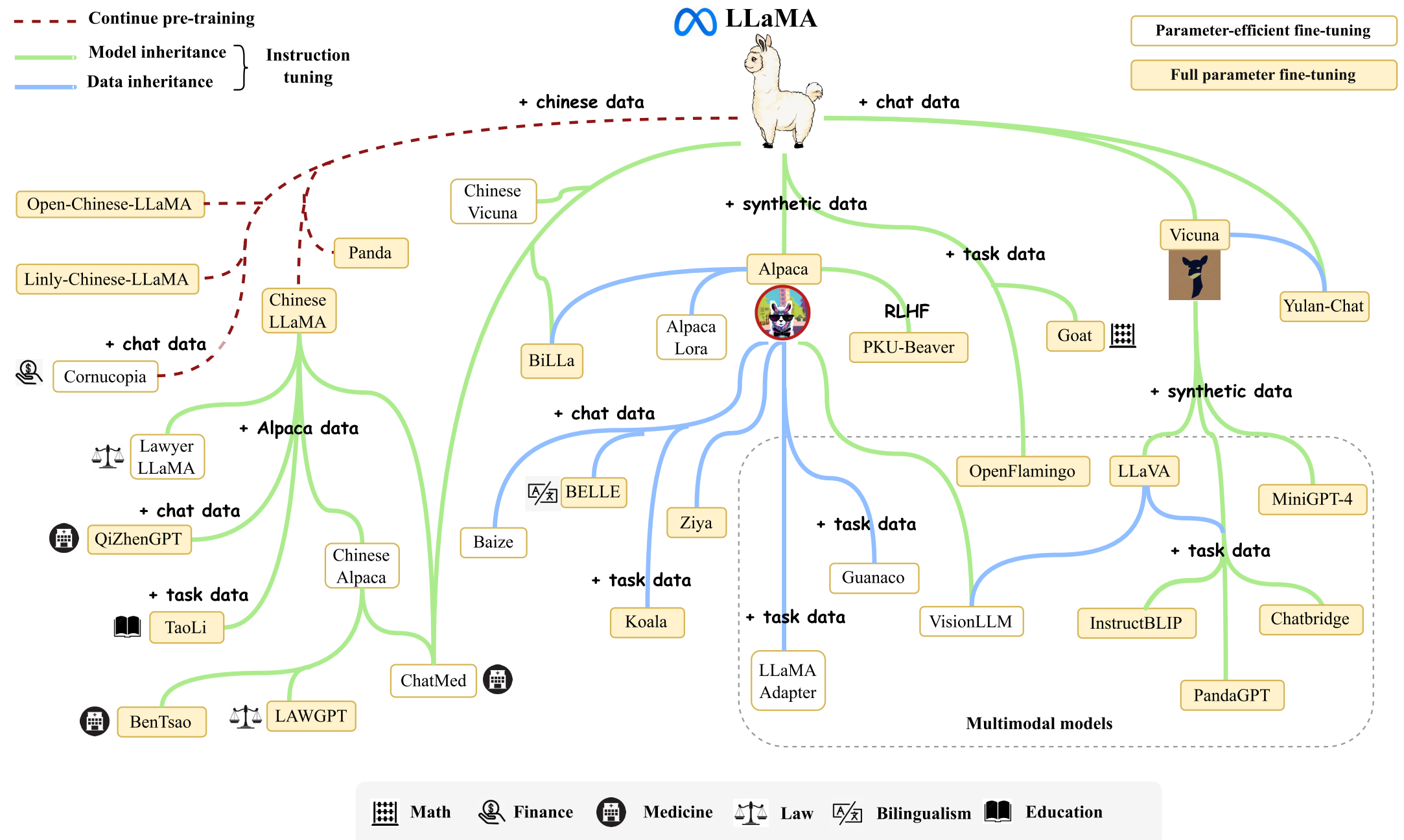} 
\caption{Overview of research work on LLaMA showing data sources, pre-training, fine-tuning and instruction tuning (image from \cite{zhao_survey_2023})}
\label{fig:llama}
\end{center}
\end{figure}

\glspl*{llm} are pre-trained on vasts amounts of textual data using a variety of strategies and techniques, which is often followed by more specific fine-tuning of the model to suit its intended use \cite{brown2020languagemodelsfewshotlearners}. Interestingly, \gls*{llama} was trained exclusively on publicly available data sources \cite{touvron2023llamaopenefficientfoundation} (Figure~\ref{fig:llama}).

The general trend has been to train ever larger models because large pre-trained Transformer models were found to be capable of performing tasks for which they had not been specifically trained on \cite{wang_what_2022}. Conversely, Hoffman et al.  \cite{Hoffmann2022} found that training a smaller model on more data for a given compute budget is more performant than only increasing model size while keeping the size of the training data unchanged. However, focussing on the optimal combination of model size and training dataset size does not take into account the importance of the speed of inference \cite{touvron2023llamaopenefficientfoundation}.  Instead Touvron et al. \cite{touvron2023llamaopenefficientfoundation} concluded that training smaller models for longer results in faster inference.

\begin{figure}[htbp]
\begin{center}
\includegraphics[width=0.55 \linewidth]{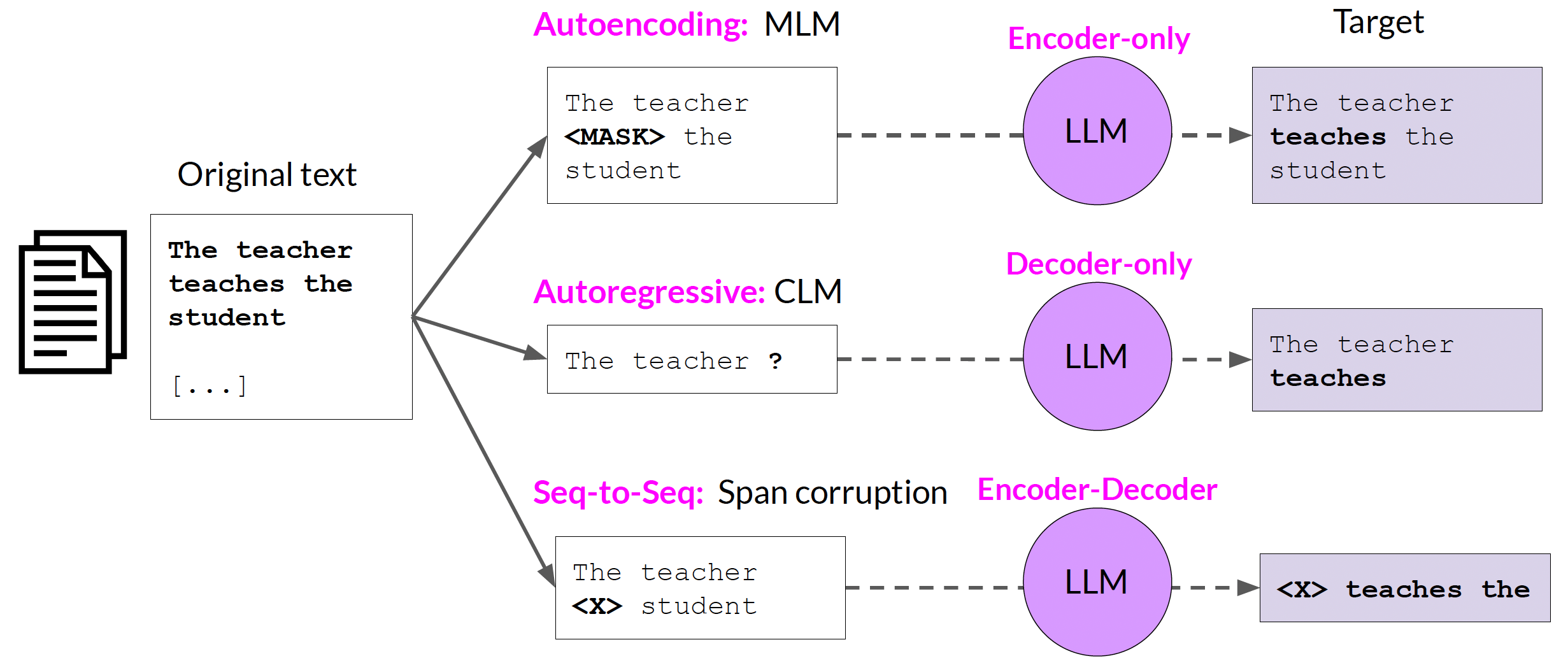} 
\caption{Model architecture and pre-training objectives (image by DeepLearning.AI \cite{barth_generative_nodate})}
\label{fig:archobjectives}
\end{center}
\end{figure}

Pre-training of \glspl*{llm} is performed using data labels. Adding labels to data prior to training (supervised learning) typically requires human annotation which is infeasible when training on very large corpora. Supervised learning for \glspl*{llm} is more typically used when training a model for a specific task, such as fine-tuning a model. Unsupervised learning on the other hand is a well known machine learning technique to learn patterns and structure from unlabelled data through methods like clustering and dimensionality reduction. \gls*{ssl} may be viewed as a `blend' of supervised and unsupervised learning. During \gls*{ssl} the unlabelled data itself provides the supervision by generating labels from the input data and this can be done in several ways, as discussed below \cite{balestriero2023}.

\begin{figure}[htbp]
\begin{center}
\includegraphics[width=0.5 \linewidth]{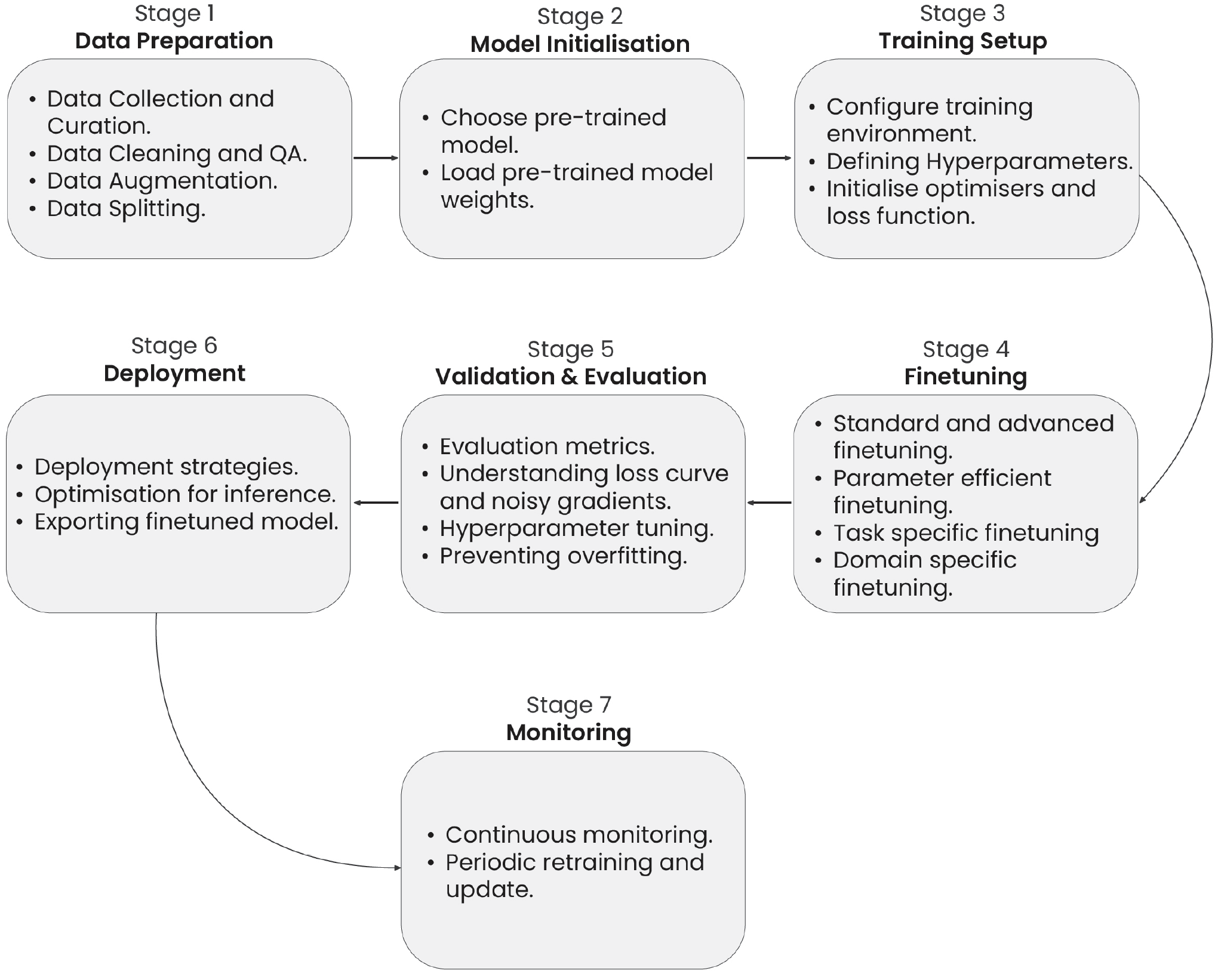} 
\caption{Pre-training and fine-tuning Large Language Models (LLMs), illustrating the seven essential stages (image from \cite{parthasarathy2024})}
\label{fig:trainingsetup}
\end{center}
\end{figure}

\subsubsection{Self-supervised learning}
\gls*{ssl} is the most popular machine learning technique for \glspl*{llm}.  This approach is often referred to as the ``dark matter of intelligence'' \cite{balestriero2023} and includes learning methods such as \gls*{clm}, \gls*{mlm}, Span-Level Masking, and Contrastive Learning, where models learn without the need to explicitly apply external labels to the data.

ELECTRA, also an encoder-only mode like BERT uses an alternative pre-training method to \gls*{mlm} called ``replaced token detection'' \cite{clark2020electra}. 

Various \gls*{ssl} approaches have been used in training foundational \glspl*{llm} (Table~\ref{tbl:ssl} on page \pageref{tbl:ssl}). Moreover, as can be seen from summary table~\ref{tbl:ssl}, page~\pageref{tbl:ssl}, some notable LLMs combine multiple self-supervised approaches to leverage the strengths of each method, e.g. BART, BERT and T5 \cite{raffel_exploring_2023, lewis2019bartdenoisingsequencetosequencepretraining}.

\begin{table}[htp]
\caption{Summary of self-supervised learning (SSL) approaches}
{\footnotesize
\begin{center} 

\begin{tabularx}{0.8\textwidth}{
		>{\raggedright\arraybackslash}p{3.74cm}
       	 	>{\raggedright\arraybackslash}X
        		>{\raggedright\arraybackslash}p{3.2cm} } 
	\toprule
	\textbf{Approach} & \textbf{Description} & \textbf{Notable Models} \\[0.4ex] 
	\midrule
Masked Language Modeling (MLM) - Figs.~\ref{fig:encoder}~\&~\ref{fig:archobjectives} & Predict masked tokens in input sequence (bi-directional) \cite{micheletti2024} & BERT, RoBERTa, ALBERT\\
\addlinespace
Causal Language Modeling (CLM) - Figs.~\ref{fig:decoder}~\&~\ref{fig:archobjectives} & Predict next token from previous tokens (uni-directional) & GPT series, Transformer-XL \\
\addlinespace
Permuted Language Modeling & Predict tokens in random order \cite{micheletti2024} & XLNet \\
\addlinespace
Next Sentence Prediction (NSP) & Predict if one sentence follows another & BERT\\
\addlinespace
Sentence Order Prediction (SOP) & Predict correct order of sentence pairs & ALBERT\\
\addlinespace
Span-Based Masking - Fig.~\ref{fig:archobjectives} & Predict missing spans of tokens & T5, BART\\
\addlinespace
Denoising Autoencoders & Reconstruct original text from corrupted input & T5, BART\\
\addlinespace
Contrastive Learning & Differentiate similar and dissimilar inputs & SimCSE\\
& & \\
	\bottomrule
	\end{tabularx}
	\end{center}
}	
\label{tbl:ssl}
\end{table}%

\subsection{Adapting \glspl*{llm} for Specific Use Cases}
\label{sec:usecases}

Large pre-trained Transformer models were found to be capable of performing tasks for which they had not been specifically trained on \cite{wang_what_2022}. This is known as ``zero-shot''  inference \cite{wang_what_2022}. However, when the output from the \gls*{llm} for a certain task is less than satisfactory, there are two main techniques to achieve better results: in-context learning and fine-tuning.

\subsubsection{In-context learning}
\Gls*{icl} refers to the capability of pre-trained \glspl*{llm} to perform new tasks by leveraging information provided within the context window, without any explicit parameter updates or fine-tuning \cite{brown2020languagemodelsfewshotlearners}. 

\begin{figure}[htbp]
\begin{center}
\includegraphics[width=0.45 \linewidth]{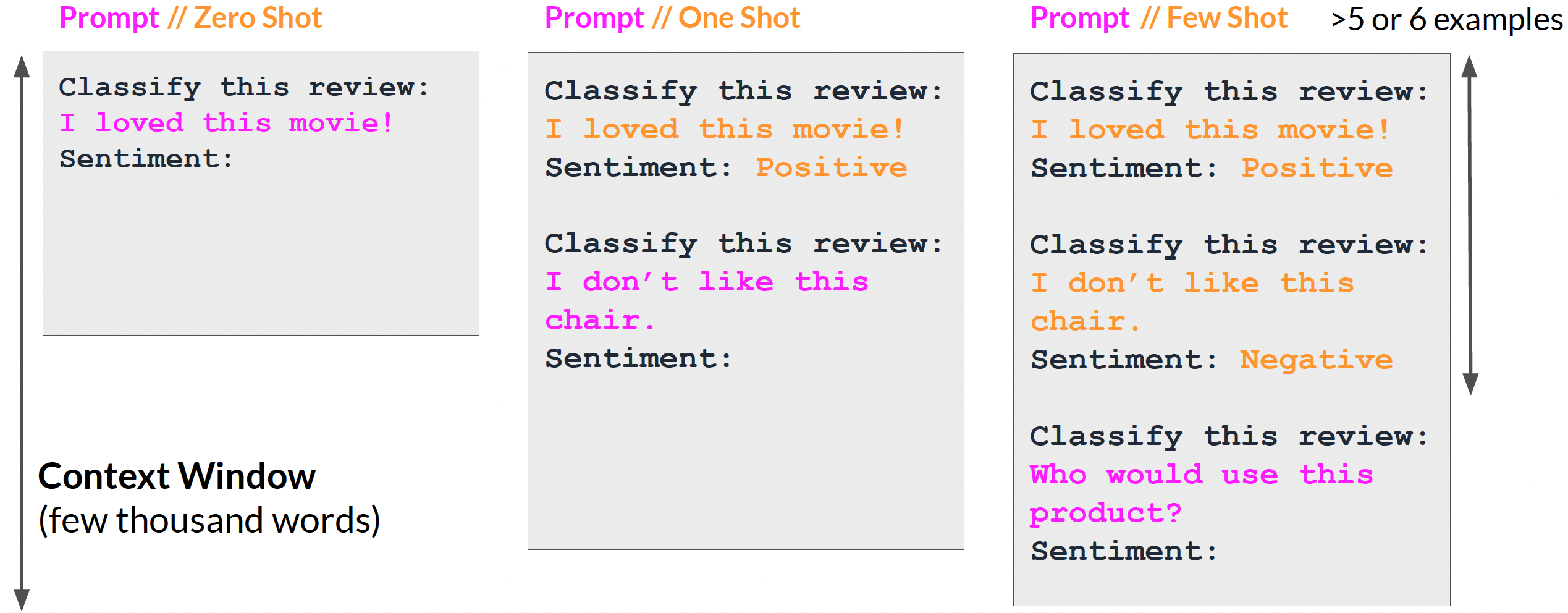} 
\caption{Example of in-context learning (ICL) (image by DeepLearning.AI \cite{ng_deeplearningai_nodate})}
\label{fig:icl}
\end{center}
\end{figure}

Instead of adjusting weights through gradient descent, the model adapts its behaviour based on examples, instructions, or demonstrations included in the prompt \cite{wei2023chainofthoughtpromptingelicitsreasoning}. Figure~\ref{fig:icl} (page~\pageref{fig:icl}) shows prompting with none, one and two examples in the context window. This strategy allows \glspl*{llm} to generalise to a wide range of tasks using natural language interactions \cite{brown2020languagemodelsfewshotlearners}.

\begin{figure}[htbp]
\begin{center}
\includegraphics[width=0.55\linewidth]{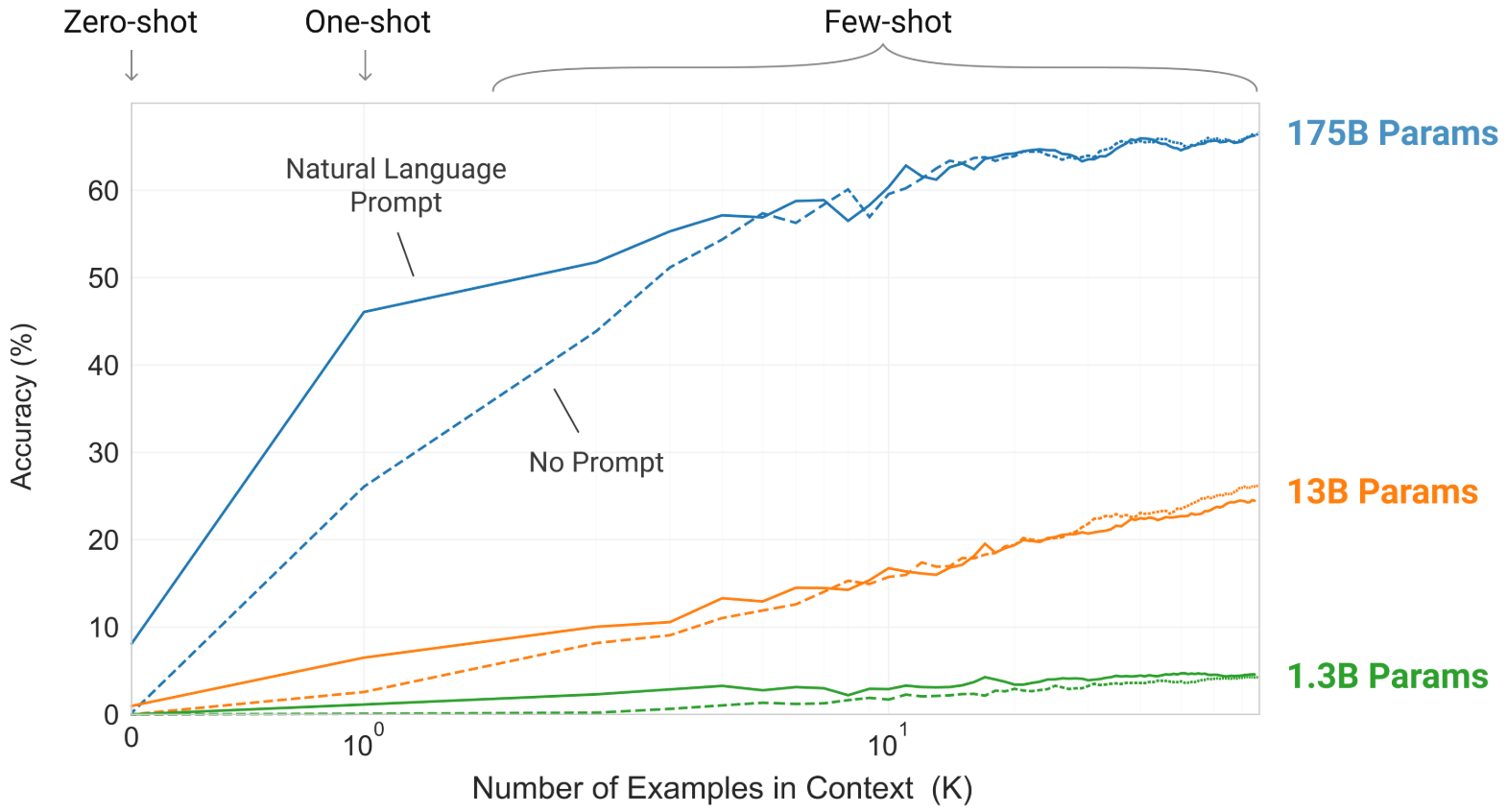}
\caption{In-context learning performance with different model sizes and number of examples \cite{brown2020languagemodelsfewshotlearners}}
\label{fig:fewshot}
\end{center}
\end{figure}

In few-shot prompting the model uses the examples provided in the context window to infer the task's structure and apply it to new inputs. The study by \cite{brown2020languagemodelsfewshotlearners} (Figure~\ref{fig:fewshot}, page~\pageref{fig:fewshot}) shows in-context learning curves with few-shot learning of a simple task. We can observe that model performance improves with increases in both model size and number of examples in the context window \cite{brown2020languagemodelsfewshotlearners}.

\begin{figure}[htbp]
\begin{center}
\includegraphics[width=0.45\linewidth]{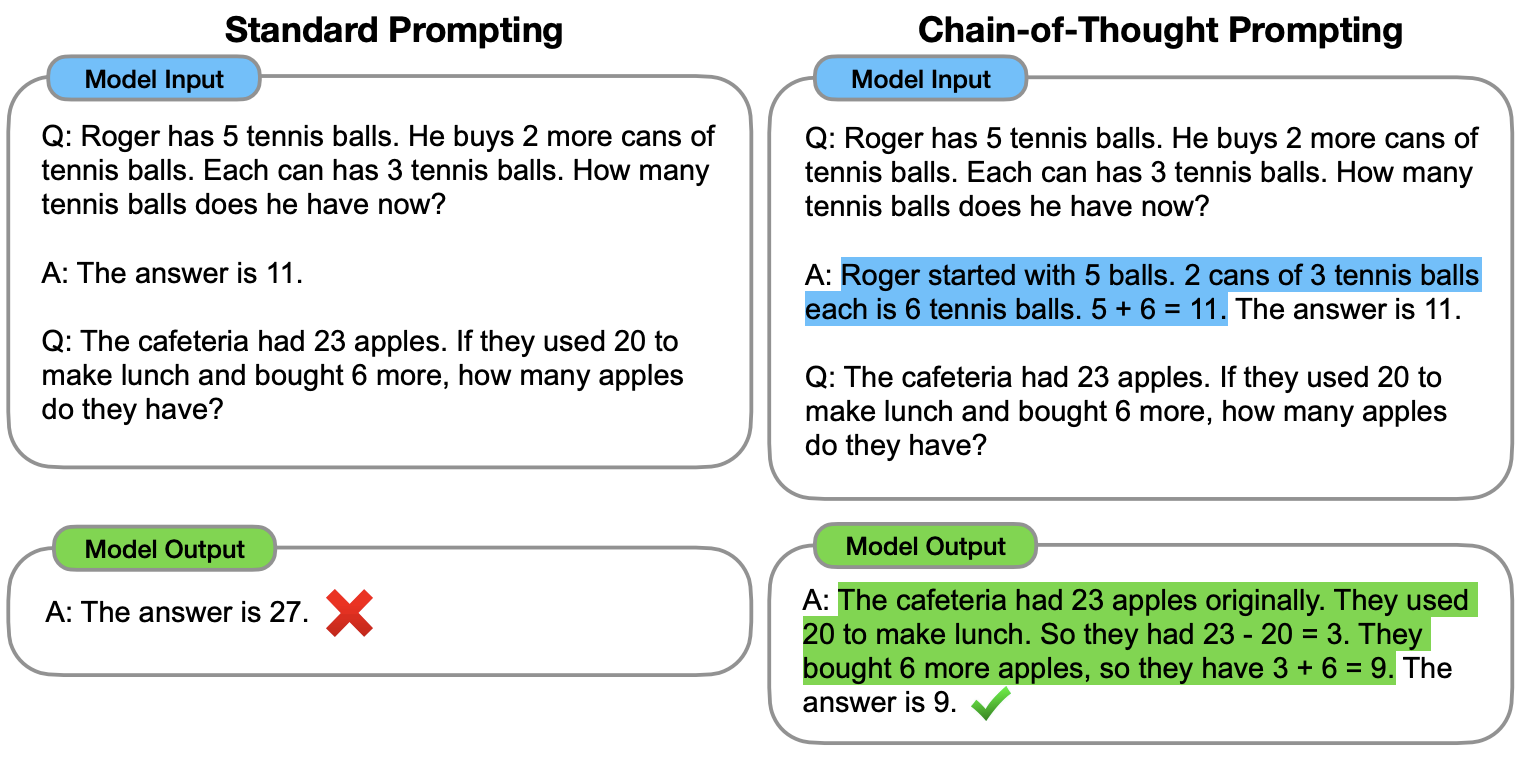}
\caption{Chain-of-thought prompting example \cite{wei2023chainofthoughtpromptingelicitsreasoning}}
\label{fig:chainofthought}
\end{center}
\end{figure}

Chain-of thought-prompting \cite{wei2023chainofthoughtpromptingelicitsreasoning} is another effective \gls*{icl} technique to help \glspl*{llm}
perform complex reasoning required for tasks such as arithmetic computations that \glspl*{llm} have been known to struggle with. In chain-of-thought reasoning, the user provides an example with the steps a human would take to achieve the desired outcome or calculation (Figure \ref{fig:chainofthought}, page \pageref{fig:chainofthought}). This technique can also be used for  commonsense and symbolic reasoning tasks \cite{wei2023chainofthoughtpromptingelicitsreasoning}.

\subsubsection{Fine-tuning}
\label{sec:finetune}
Fine-tuning is the process of updating pre-trained \gls*{llm} weights by training on specific datasets for chosen tasks \cite{brown2020languagemodelsfewshotlearners, raffel_exploring_2023}. \Gls*{sft} \cite{devlin2019bertpretrainingdeepbidirectional}, \gls*{rlhf} \cite{ouyang2022}, and \gls*{peft} \cite{Houlsby2019} are three of the most popular fine-tuning approaches for LLMs. 

\begin{figure}[htbp]
\begin{center}
\scalebox{0.75} {
\begin{tikzpicture}[mindmap, grow cyclic, text width=2.5cm, align=flush center,
  every node/.style={concept, font=\footnotesize}, concept color=blue!40,
  level 1/.append style={level distance=4cm, sibling angle=120},
  level 2/.append style={level distance=3cm, sibling angle=45}]

\node{\textbf{Fine-Tuning Techniques}}
    child[concept color=green!40] { node {\textbf{1. Full Model Fine-Tuning}}
        child { node {Supervised\\Fine-Tuning (SFT)}}
        child { node {Instruction\\Fine-Tuning}}
        child { node {RLHF}}
        child { node {Multi-Task\\fine-tuning}}
        child { node {Continual\\Learning}}
    }
    child[concept color=red!40] { node {\textbf{2. Parameter-Efficient Fine-Tuning}}
        child { node {Adapter\\Layers}}
        child { node {Low-Rank\\Adaptation\\(LoRA)}}
        child { node {Prefix and\\Prompt tuning}}
        child { node {Sparse\\fine-tuning}}
        child { node {Frozen Layers}}
    }
    child[concept color=yellow!10!brown] { node {\textbf{3. Model Compression and Deployment Optimization}}
        child { node {Knowledge\\Distillation}}
        child { node {Quanti-\\sation-\\Aware}}
        child { node {Progressive\\Training}}
        child { node {Federated\\Learning}}
    };
\end{tikzpicture}
}
\caption{LLM fine-tuning categories}
\label{fig:finetune}
\end{center}
\end{figure}
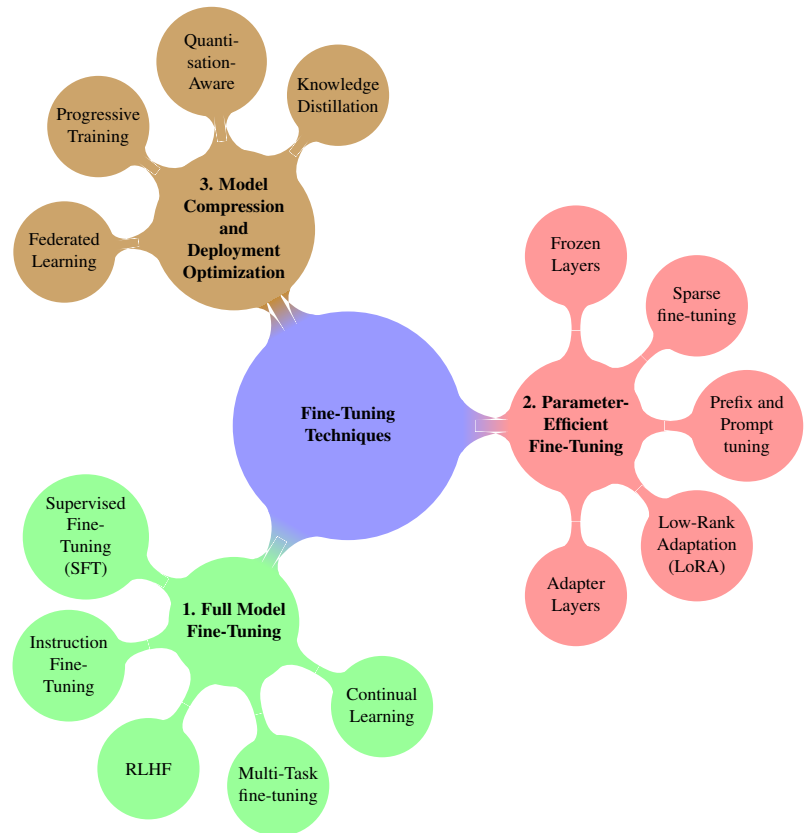

However, there are several other fine-tuning approaches that can be employed. They can broadly be categorised as full model fine-tuning (e.g. \gls*{sft} \cite{devlin2019bertpretrainingdeepbidirectional} and \gls*{rlhf}  \cite{ouyang2022}), \gls*{peft} (e.g. \gls*{lora} \cite{hu_lora_2021}), and model compression and deployment optimisation (e.g. quantisation-aware fine-tuning as used for Q8BERT LLM \cite{zafrir2019}). A visualisation of this grouping and associated fine-tuning techniques in each category are shown as a mind map  in Figure~\ref{fig:finetune} on page~\pageref{fig:finetune}. 

Sometimes a single technique may be insufficient in delivering the desired outcomes, and instead we can combine multiple fine-tuning strategies, leveraging the strengths of each technique in order to address shortcomings, such as solving multiple constraints simultaneously or the need to optimise for performance, efficiency, and alignment. For example, combining \gls*{rlhf} with \gls*{lora} would yield models that are both aligned with human preferences and parameter-efficient. 
\clearpage        

To complement the visual representation of fine-tuning strategies in Figure~\ref{fig:finetune} on page~\pageref{fig:finetune}, Table~\ref{tbl:finetune} on page~\pageref{tbl:finetune} provides a bit more detail for each of the strategies in the diagram. 

{\footnotesize
\begin{center} 
\begin{longtable}{>{\raggedright\arraybackslash}p{3.7cm} 
			  >{\raggedright\arraybackslash}p{5 cm} 
			  >{\raggedright\arraybackslash}p{5 cm} }
\caption{Summary of fine-tuning strategies} \\
\toprule
\textbf{Fine-Tuning Approach} & \textbf{Description} & \textbf{When to Use}  \\
\midrule
\endfirsthead

\toprule
\textbf{Fine-Tuning Approach} & \textbf{Description} & \textbf{When to Use}  \\
\midrule
\endhead

\midrule
\multicolumn{3}{r}{\small\itshape Continued on next page} \\
\bottomrule
\endfoot

\bottomrule
\endlastfoot

\textbf{Supervised Fine-Tuning (SFT)} &
Updating all pre-trained model parameters on labeled data for a specific task \cite{brown2020languagemodelsfewshotlearners}. &
\vspace{-10pt}
\begin{itemize} [topsep=0pt, leftmargin=*, labelindent=0pt, noitemsep,]
    \item Substantial amount of labelled data available.
    \item Well-defined tasks requiring high accuracy.
\end{itemize} \\

\addlinespace

\textbf{Instruction Fine-Tuning} &
Using  instruction-response pairs to enhance the model's ability to follow human instructions \cite{wei2022zeroshot}. &
\vspace{-10pt}
\begin{itemize} [topsep=0pt, leftmargin=*, labelindent=0pt, noitemsep,]
    \item Improving the model's ability to understand and execute human instructions.
    \item Developing assistant-like applications.
\end{itemize}  \\

\addlinespace

\textbf{Reinforcement Learning from Human Feedback (RLHF)} &
Using human feedback to train a reward model to guide models via reinforcement learning to align with human preferences \cite{ouyang2022} .&
\vspace{-10pt}
\begin{itemize} [topsep=0pt, leftmargin=*, labelindent=0pt, noitemsep,]
    \item Aligning model outputs with human values and preferences.
    \item Improving response quality and safety.
\end{itemize} \\

\addlinespace
\textbf{Multi-Task Fine-Tuning} &
Simultaneously fine-tuning for multiple tasks to achieve better generalisation \cite{ning2024}. &
\vspace{-10pt}
\begin{itemize} [topsep=0pt, leftmargin=*, labelindent=0pt, noitemsep,]
    \item Models that perform well on multiple tasks.
    \item To improve generalisation.
\end{itemize} \\

\addlinespace

\textbf{Continual Learning} &
Sequentially fine-tuning the model on new tasks while preserving previous knowledge \cite{kirkpatrick_overcoming_2017}.
&
\vspace{-10pt}
\begin{itemize} [topsep=0pt, leftmargin=*, labelindent=0pt, noitemsep,]
    \item Model needs to adapt over time, e.g. evolving data distributions.
    \item To prevent catastrophic forgetting.
\end{itemize} \\

\addlinespace

\textbf{Fine-Tuning with Frozen Layers} &
Freezing certain layers and updating only the top layers to reduce computation and retain general knowledge \cite{Panoutsopoulos2024} .
&
\vspace{-10pt}
\begin{itemize} [topsep=0pt, leftmargin=*, labelindent=0pt, noitemsep,]
    \item Limited computational resources.
    \item Limited fine-tuning data available.
    \item To prevent overfitting.
\end{itemize} \\

\addlinespace
\textbf{Sparse Fine-Tuning} &
Updating only a subset of model parameters 
relevant to the new tasks \cite{mu2024}.  &
\vspace{-10pt}
\begin{itemize} [topsep=0pt, leftmargin=*, labelindent=0pt, noitemsep,]
    \item For computational efficiency.
    \item Limited fine-tuning data available.
    \item To prevent overfitting.
\end{itemize} \\

\addlinespace

\textbf{Prefix Tuning and Prompt Tuning} &
Adding trainable continuous prompts or prefix tokens to inputs to adapt model with minimal changes to original parameters \cite{li2021prefixtuning, shrestha_post-translational_2024}. 
&
\vspace{-10pt}
\begin{itemize} [topsep=0pt, leftmargin=*, labelindent=0pt, noitemsep,]
    \item For parameter-efficient fine-tuning.
    \item Adapting to multiple tasks with minimum alteration to core weights.
\end{itemize} \\

\addlinespace
\textbf{Low-Rank Adaptation (LoRA)} &
LoRA freezes model weights and inserts trainable low-rank matrices in the model layers which reduces the number of trainable parameters \cite{hu_lora_2021}.&
\vspace{-10pt}
\begin{itemize} [topsep=0pt, leftmargin=*, labelindent=0pt, noitemsep,]
    \item Limited computational resources.
    \item Rapid experimentation required.
\end{itemize}  \\

\addlinespace

\textbf{Adapter Layers} &
Inserting lightweight adapter modules in model layers to adapt to new tasks, only updating adapter parameters \cite{Houlsby2019, lialin_scaling_2023} .&
\vspace{-10pt}
\begin{itemize} [topsep=0pt, leftmargin=*, labelindent=0pt, noitemsep,]
    \item Limited computational resources.
    \item For multi-task learning with a shared base model .
    \item Avoiding catastrophic forgetting because base model is unchanged.
\end{itemize} \\

\addlinespace

\textbf{Federated Learning for LLMs} &
Fine-tuning across decentralised data sources while preserving privacy \cite{mcmahan2023}. &
\vspace{-10pt}
\begin{itemize} [topsep=0pt, leftmargin=*, labelindent=0pt, noitemsep,]
    \item Data privacy is a concern.
    \item Well suited for sensitive or proprietary data.
\end{itemize} \\

\addlinespace

\textbf{Progressive Training} &
Trains models in stages, starting with smaller models and gradually increasing complexity \cite{lu2024yoda}. &
\vspace{-10pt}
\begin{itemize} [topsep=0pt, leftmargin=*, labelindent=0pt, noitemsep,]
    \item Large datasets or models.
    \item Improve generalisation over progressive complexity.
\end{itemize} \\

\addlinespace

\textbf{Quantisation-Aware Fine-Tuning} &
Simulating quantisation effects to ensure robustness and maintain performance on low-precision hardware (e.g., 8-bit systems) \cite{zafrir2019}. &
\vspace{-10pt}
\begin{itemize} [topsep=0pt, leftmargin=*, labelindent=0pt, noitemsep,]
    \item Deploying on devices with limited computational power.
    \item Reducing model size and increasing inference speed.
\end{itemize} \\

\addlinespace

\textbf{Knowledge Distillation} &
Training a smaller model to mimic a larger model for deployment in resource-constrained environments \cite{hinton2015}. &
\vspace{-10pt}
\begin{itemize} [topsep=0pt, leftmargin=*, labelindent=0pt, noitemsep,]
    \item A lightweight model is required.
    \item Ideal for real-time inference.
    \item To compress models without significant performance loss.
\end{itemize} \\

& & \\
\bottomrule
\label{tbl:finetune}
\end{longtable}%
\end{center}
}	

\subsection{Creating a bespoke LLM }
\label{sec:bespoke}
In some instances it may be preferable to develop a bespoke \gls*{llm} instead of fine-tuning one of the popular foundational models. To do this, the following steps provide a general approach:

\begin{enumerate}
\item \textit{Data Selection and Preparation}
\begin{itemize}
\item \textit{Data gathering}: The foundation of any LLM is the data it learns from. Therefore, identifying the appropriate and relevant data sources is an important first step in developing a bespoke \gls*{llm} and requires a clear understanding of the key objectives of the LLM. The data gathering exercise typically involves obtaining extensive text data from various sources such as books, websites, and articles, but in other cases the inclusion of a highly diverse corpus of text data may be less relevant and attention is instead focussed on sourcing only a few, but high quality datasets to train the model on. Nonetheless, some additional refinements may be achieved at a later stage by employing a variety of learning approaches as discussed in Section~\ref{sec:usecases}~\nameref{sec:usecases}.

\item \textit{Preprocessing}: The data typically needs some degree of preprocessing, such as data cleansing to remove noise and irrelevant content, normalisation to standardise text formats, and tokenisation to convert text into a format that the model can understand. 

\item \textit{Annotation}: If supervised learning is involved, this stage may also include annotating the data with labels. 

\item \textit{Training data}: Finally, the dataset is split into training, validation, and test datasets to enable effective learning and unbiased evaluation. The training data allocation is usually set around 15\%.
\end{itemize}

\item \textit{Model Design and Configuration}\\
Choosing the right model architecture is critical to achieving the desired performance. For LLMs, Transformer-based architectures are commonly used due to their ability to capture long-range dependencies in text. This step involves configuring the model’s parameters, such as the number of layers, hidden units, and attention heads, to balance performance with computational feasibility. Hyperparameter tuning is conducted to find optimal settings for learning rate, batch size, and regularisation techniques, which can significantly impact the efficiency and effectiveness of the training process.
\item \emph{Training the model} \\
Once the data has been sourced and cleaned, and the model architecture chosen, the training environment needs to be set up. This includes selecting appropriate loss functions (like cross-entropy loss) and optimisers (such as \href{https://keras.io/api/optimizers/adam/}{Adam} or \href{https://keras.io/api/optimizers/adafactor/}{Adafactor} \cite{keras_2024}). The model learns by minimising the loss function over the training data, adjusting its internal parameters to improve predictions. Throughout training, it is important to monitor metrics like loss and accuracy, and to validate the model on the validation set to prevent overfitting, a problem inherent in machine learning techniques. 
\item \emph{Fine-tuning and Deployment}\\
Once a model is trained we need to evaluate it against expected behaviour and through approaches such as fine-tuning and prompt engineering to ensure that the model performs as desired. Using the prior technique will adjust model weights, whereas the latter leaves the original weights in tact. Further actions such as developing APIs to access the trained LLM may then be undertaken and deployed in production. Post deployment it is crucial to be cognisant of ethical implications and legal considerations, including assessment of unintended biases. 
\end{enumerate}

Ultimately LLM development is an iterative process and by leveraging information such as user feedback, metrics of loss and accuracy, changes to task requirements and/or current data, and compliance with the validation set to prevent overfitting, we can ensure that the LLM remains relevant throughout its life.  

\subsection{Interacting with LLMs}
\label{sec:interact}
There is a myriad of ways in which we can interact with \glspl*{llm}, depending on the desired end goal(s).  Some interactions such as fine-tuning (See Section~\ref{sec:finetune}, page~\pageref{sec:finetune}) adjust the base model while others focus on the most effective way to perform tasks and extract information without adjusting the underlying model. Figure~\ref{fig:finetune}, page~\pageref{fig:finetune} visually summarises the various approaches.

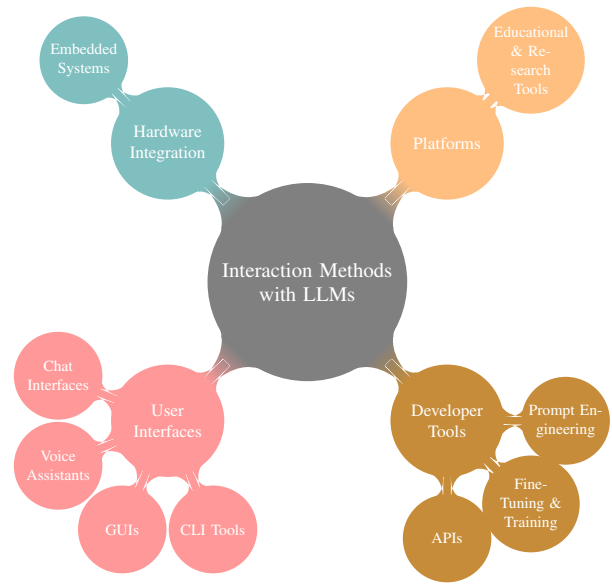
\begin{figure}[htbp]
\begin{center}
\scalebox{0.65} {
\begin{tikzpicture} [scale=0.8]
\path[mindmap, concept color=gray, text=white]
node[concept] {Interaction\ Methods\ with LLMs}
[
    grow cyclic,
    level 1/.append style={level distance=5cm, sibling angle=90},
    level 2/.append style={level distance=3cm, sibling angle=45}
]
child[concept color=red!40] {
    node[concept] {User Interfaces}
    child { node[concept] {Chat Interfaces} }
    child { node[concept] {Voice Assistants} }
    child { node[concept] {GUIs} }
    child { node[concept] {CLI Tools} }
}
child[concept color=brown!90!yellow] {
    node[concept] {Developer Tools}
    child { node[concept] {APIs} }
    child { node[concept] {Fine-Tuning \& Training} }
    child { node[concept] {Prompt Engineering} }
}
child[concept color=orange!50] {
    node[concept] {Platforms}
    child { node[concept] {Educational \& Research Tools} }
}
child[concept color=teal!50] {
    node[concept] {Hardware Integration}
    child { node[concept] {Embedded Systems} }
};
\end{tikzpicture}
}
\caption{Ways of interacting with an LLM}
\label{fig:interaction}
\end{center}
\end{figure}

Table~\ref{tbl:interaction}, page~\pageref{tbl:interaction}, gives a brief overview of these methods, but arguably the most common and well known way of interacting with \glspl*{llm} is through a chat bot such as \href{https://chat.openai.com/}{ChatGPT (by OpenAI)}, \href{https://bard.google.com/}{BARD (by Google)}, \href{https://www.anthropic.com/}{Claude (by Anthropic)} and \href{https://www.bing.com/chat}{Bing Chat (by Microsoft, powered by OpenAI)} \cite{biever_chatgpt_2023, Gabriel2023, hao-wen_challenges_2023}.

\begin{table}[h!]
\caption{Methods to Interact with LLMs}
\centering
\begin{tabular}{@{}lp{10cm}@{}}
\toprule
\textbf{Method} & \textbf{Explanation} \\
\midrule
Chat Interfaces & User-friendly platforms for real-time conversational interaction with LLMs. \\
\addlinespace
Voice Assistants & Use of speech to interact with \glspl*{llm} in voice-enabled applications. \\
\addlinespace
Graphical User Interfaces (GUIs) & GUI-based applications enabling interaction with \glspl*{llm} without coding. \\
\addlinespace
Command-Line Interfaces (CLIs) & Interaction with \glspl*{llm} via command-line tools for scripting and automation tasks. \\
\addlinespace
APIs \& Wrappers & Programmatic access to \glspl*{llm} and associated libraries for ease of integration into applications and services. \\
\addlinespace
Fine-Tuning and Training & Adjusting model parameters to perform specialised tasks using machine learning tools. \\
\addlinespace
Prompt Engineering & The art of crafting specific prompts to elicit desired outputs by \glspl*{llm}.  \\
\addlinespace
Educational and Research Tools & Using platforms such as Jupyter and Colab for experimenting and learning with \glspl*{llm}. \\
\addlinespace
Embedded Systems & Integration of \glspl*{llm} into hardware devices for natural language understanding. \\

\bottomrule
\end{tabular}
\label{tbl:interaction}
\end{table}

In Section~\ref{sec:prompt}, page~\pageref{sec:prompt}, we discuss prompt engineering in more detail, a simple and effective way for most users to harness the knowledge, and explore the functionality, of an \gls*{llm}. However, a flexible, powerful and effective way of interacting with \glspl*{llm} is through \glspl*{api}, but that requires a higher level of technical expertise. Several of the well known pre-trained \glspl*{llm} provide \glspl*{api}, some are open source and others not. The more popular \glspl*{api} are: \href{https://huggingface.co/docs/api-inference/en/index}{Hugging Face's Transformers Library and Inference \gls*{api}}, \href{https://cloud.google.com/natural-language}{Google Cloud's Natural Language \gls*{api}}, \href{https://www.ibm.com/products/natural-language-processing}{IBM Watson Language Translator \gls*{api}}, \glspl*{api} to access \gls*{bert} can be obtained via \href{https://github.com/google-research/bert}{Google Research \gls*{bert} repository} or through \href{https://huggingface.co/bert-base-uncased}{Hugging Face's \gls*{bert} model webpage}, and \glspl*{api} for \href{https://platform.openai.com/docs/overview}{OpenAI GPT-4o and GPT-4o mini}.

\subsubsection{Prompt Engineering}
\label{sec:prompt}
Prompt engineering is a technique used to maximise the effectiveness of an existing LLM without altering its internal structure. The process comprises three parts: the \emph{prompt} itself is the model input, model \emph{inference} is the generation of text in response to the prompt, and lastly \emph{completion} is the resulting output text.  The \emph{context window} is the all the text and memory that is available.
 
By carefully crafted prompts, users can harness these models more effectively, leading to better outcomes in tasks ranging from simple queries to complex problem solving, but it has limitations. One effective strategy to improve model outcomes is by including examples inside the context window  (Figure~\ref{fig:icl}, page~\pageref{fig:icl}). This process is called \emph{in-context learning} and the variations of in-context learning are: \cite{brown2020languagemodelsfewshotlearners}:
\begin{itemize}[noitemsep]
\item \emph{zero-shot} inference - no examples provided
\item \emph{one-shot} inference - one example provided
\item \emph{few-shot} inference - more than one example provided
\end{itemize}

We can also view prompt engineering as a complementary technique to fine-tuning by using it to generate training data or as an interim solution to improve the model's performance. A general guide for progressing on to fine-tuning is when the number of examples (few shot learning) is growing to more than 5 or 6, with diminishing improvements in LLM output. Nonetheless, the research study by Brown, T.B., et al. \cite{brown2020languagemodelsfewshotlearners} used a few dozen examples in their few-shot settings (Figure~\ref{fig:fewshot}, page~\pageref{fig:fewshot}).

\clearpage
\subsubsection{Summary of LLM Overview}
\label{sec:summary}
\begin{figure}[htbp]
\begin{center}
\includegraphics[width=0.75 \linewidth]{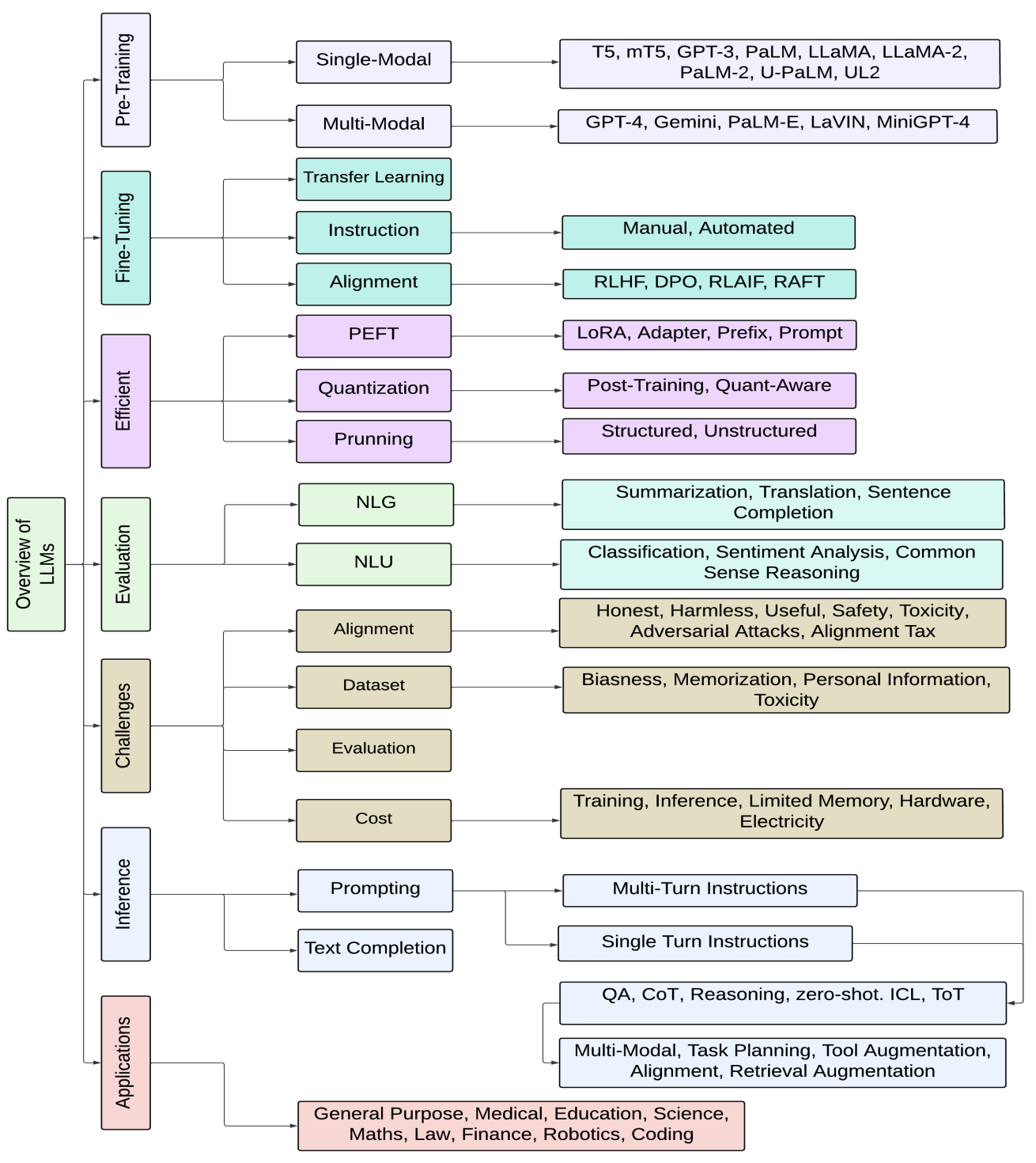} 
\caption{Overview of the various characteristics, activities and strategies of LLMs : 1. Pre-Training 2. Fine-Tuning 3. Efficient 4. Inference 5. Evaluation 6. Applications
7. Challenges (image by Naveed, H., et al. \cite{naveed2024})}
\label{fig:llmoverview}
\end{center}
\end{figure}

This overview of \glspl*{llm} is visually captured in Figure~\ref{fig:llmoverview} on page~\pageref{fig:llmoverview} depicting the different phases and characteristics of \glspl*{llm}. 

\section{LLMs Orchestrated with Other Technologies}
\label{orchestration}

Orchestration of \glspl*{llm} with traditional information retrieval systems has been explored since the early stages of this technology. Google researchers developed a platform in 2017 to speed up the creation and maintenance of production platforms when combining components of their TensorFlow machine learning system \cite{baylor2017tfx}. Some of those same techniques are present in more modern systems today. In 2018 medical informatics researchers combined image caption-generation engines with structured data stores to yield better captions \cite{li2018hybrid}. By 2020, Google and collaborators were retrieving textual data from a textual knowledge corpus based on Wikipedia documents to augment pre-training of \glspl*{llm}.

Starting in 2022, LangChain\footnote{\url{https://www.langchain.com/}} was released as an open source software project to assist software developers with the integration of \glspl*{llm} into software applications. A venture-funded company was later built around the project. Many other orchestration platforms have since appeared, including close competitor n8n\footnote{\url{https://n8n.io/}}, Haystack\footnote{\url{https://haystack.deepset.ai/}} and LlamaIndex\footnote{\url{https://www.llamaindex.ai/}}. Many of these systems are released under open source licenses.

The subfield of \gls*{kr} provides the intellectual foundation and practical tooling to represent data in ways that serve as input to other \gls*{ai} or \gls*{dms} systems. \gls*{kr} systems include ontologies, metadata, and other forms of structured information that enable meaningful representation of domain-specific knowledge. The orchestration of structured \gls*{kr}  and unstructured \gls*{llm} systems can result in an \gls*{llm} fine-tuned for a specific domain by runtime reference to a specific ontology \cite{holzinger2017, schoch_nl2ibe_2024}.  For example, application of such an orchestrated system in an engineering domain can output engineering intention artefacts \cite{schoch_nl2ibe_2024}. Another example is orchestration in a medical domain where it is highly desirable to have \gls*{xai} so that the \gls*{llm} can explain the reasoning leading to the conclusions and output making it verifiable by humans. This is especially important given the sensitive and critical nature of medical advice and the potential harmful implications of mis-diagnoses \cite{holzinger2017}. 

Alternatively, \gls*{rag} may be orchestrated with \glspl*{llm} to create custom chatbots or agents, document summarisation systems using specialist vocabularies and provide data integration with existing systems \cite{guu2020retrieval}.

\begin{figure}[htbp]
\begin{center}
\includegraphics[width=0.9\linewidth]{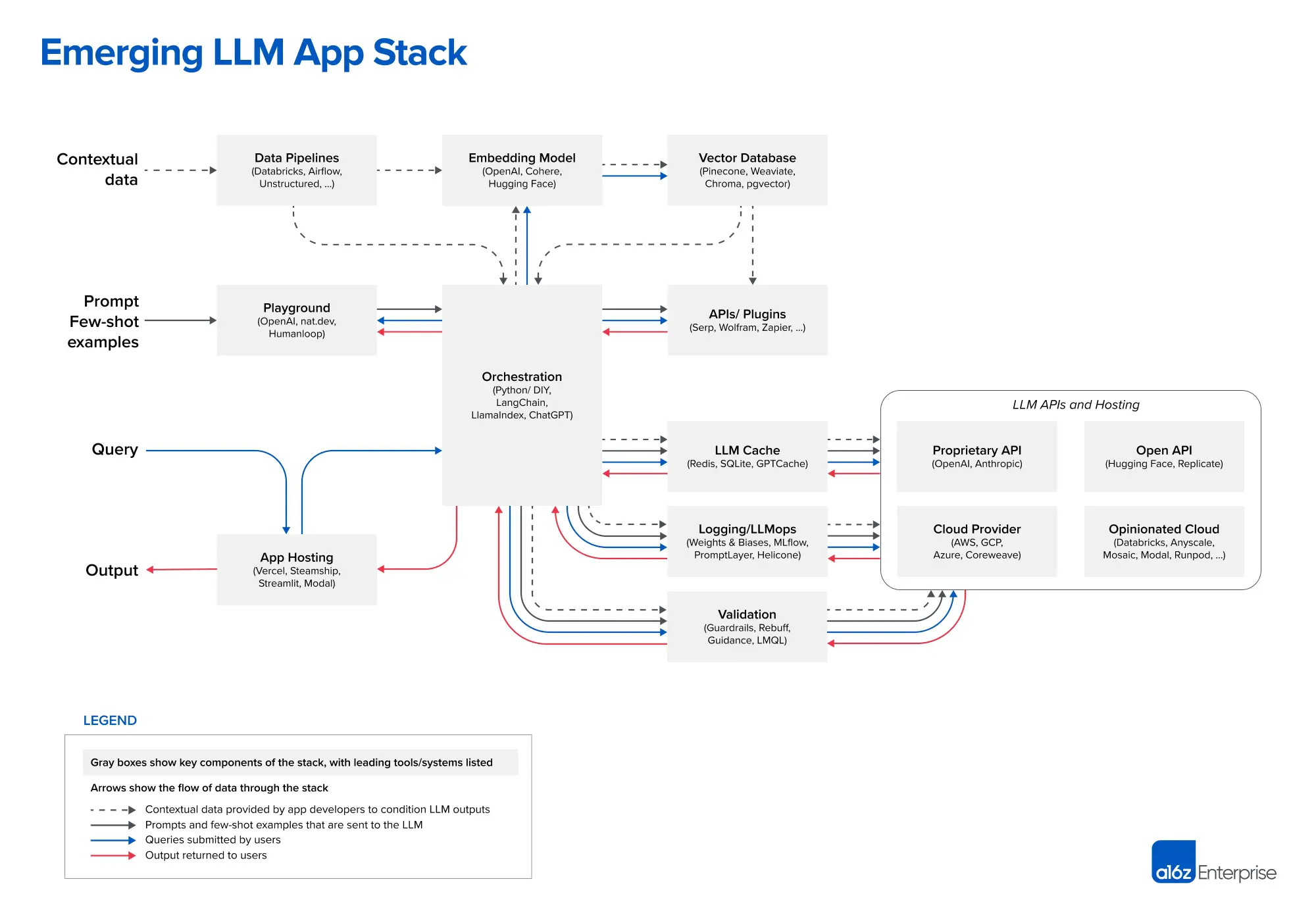}
\caption{Emerging \gls*{llm} application stack (provided by Andreessen Horowitz Enterprise)}
\label{fig:emergingorchestrationstack}
\end{center}
\end{figure}

Figure \ref{fig:emergingorchestrationstack} illustrates the emerging architecture of systems orchestrated with \glspl*{llm} in 2024. That figure is courtesy of Andreessen Horowitz Enterprise\footnote{\url{https://a16z.com/enterprise/}}.

\section{Risks and Mitigations}
\label{risksandmitigations}

\subsection{Catastrophic Forgetting}
\label{sec:forgetting}

Catastrophic forgetting, or catastrophic interference, is when neural networks, including \glspl*{llm}, become less performant on tasks that they previously excelled at \cite{luo2024empiricalstudycatastrophicforgetting, zhai_investigating_2024, li_revisiting_2024}. In other words, they essentially ``forget'' previously learned information \cite{luo2024empiricalstudycatastrophicforgetting}. This behaviour is typically observed when \glspl*{llm} are fine-tuned sequentially on different tasks or datasets, a process known as continual learning \cite{kirkpatrick_overcoming_2017}. The underlying cause is that the fine-tuning exercise updates the model's weights to optimise performance on the new task. Several strategies have been proposed to prevent catastrophic forgetting, such as:

\begin{itemize}
\item Regularisation-based method: 
Adding regularisation terms to penalise significant changes to important weights. For example \gls*{ewc} adds a regularisation term to the loss function for changes to important weights \cite{kirkpatrick_overcoming_2017}. 
\item Replay-based method:
Retraining the model on a mix of old and new data, or using synthetic data generated from the model's memory of previous tasks \cite{rolnick_experience_2019, MerlinGabriele2022PRfC}.
\item Architectural methods: 
Using separate subnetworks for different tasks, or dynamically expanding the network. For example, progressive neural networks that create new subnetworks for each task while keeping the original fixed \cite{rusu_progressive_2022}, and adapter modules that can be inserted into the network and fine-tuned separately for each task \cite{PfeifferJonas2021ANtc}.
\end{itemize}

\subsection{Model Collapse}
\label{sec:modelcollapse}

\glspl*{llm} are trained on many public data sources as described in Section~\ref{overviewofllms} \nameref{overviewofllms}. Since many users are using \glspl*{llm} to generate content that is being put onto those same public fora, future versions of those \glspl*{llm} are very likely to ingest content generated by earlier versions of themselves. It is not difficult to envision a future in which \glspl*{llm} become trained on an ever-increasing amount of machine-generated content and a decreasing amount of human-generated content. The ramifications are intriguing; without a change in the way the models are trained their weights will be increasingly influenced by machine-generated content. Human-generated content could even become a minority input for some models.

The unintended or unrecognised prevalence of machine-generated content in training data coupled with the failure of \glspl*{llm} to differentiate human- and machine-generated content is known as model collapse \cite{Shumailov_Shumaylov_Zhao_Gal_Papernot_Anderson_2024}. An \gls*{llm} in model collapse would not treat human-generated content in a preferred manner. Instead, a positive feedback loop would be set up whereby new \glspl*{llm} will learn to write like old \glspl*{llm}.

Possible mitigations for model collapse include the use of data provenance techniques to label human- and/or machine-generated content  \cite{baioumy_ai_2024}. Such approaches are limited to mitigating, not solving, the problem of model collapse because many systems and users may simply fail to provide or choose to ignore data provenance hints.

Other mitigations may be possible via governmental \gls*{ai} strategies and subsequent regulation \cite{tjondronegoro_strategic_2024}. A common analysis technique for such frameworks is the PESTEL analysis technique. PESTEL is an acronym standing for political, economic, social, technological, environmental, and legal factors in an environment external to an organisation \cite{aguilar1967scanning}. Tjondronegoro notes that a PESTEL analysis of \gls*{ai} adoption barriers and themes suggests that ``Data availability, quality, and structure'' fall under the technology rubric \cite{tjondronegoro_strategic_2024}. Governments may choose to selectively regulate some data availability to reduce negative consequences of model collapse.

None of the currently-identified mitigations to model collapse appear to be sufficient to prevent the phenomenon from occurring. More research into this area is urgently needed.

\subsection{Jailbreak Attacks}
\label{sec:jailbreak}

\begin{figure}[htbp]
\begin{center}
\includegraphics[width=0.7\linewidth]{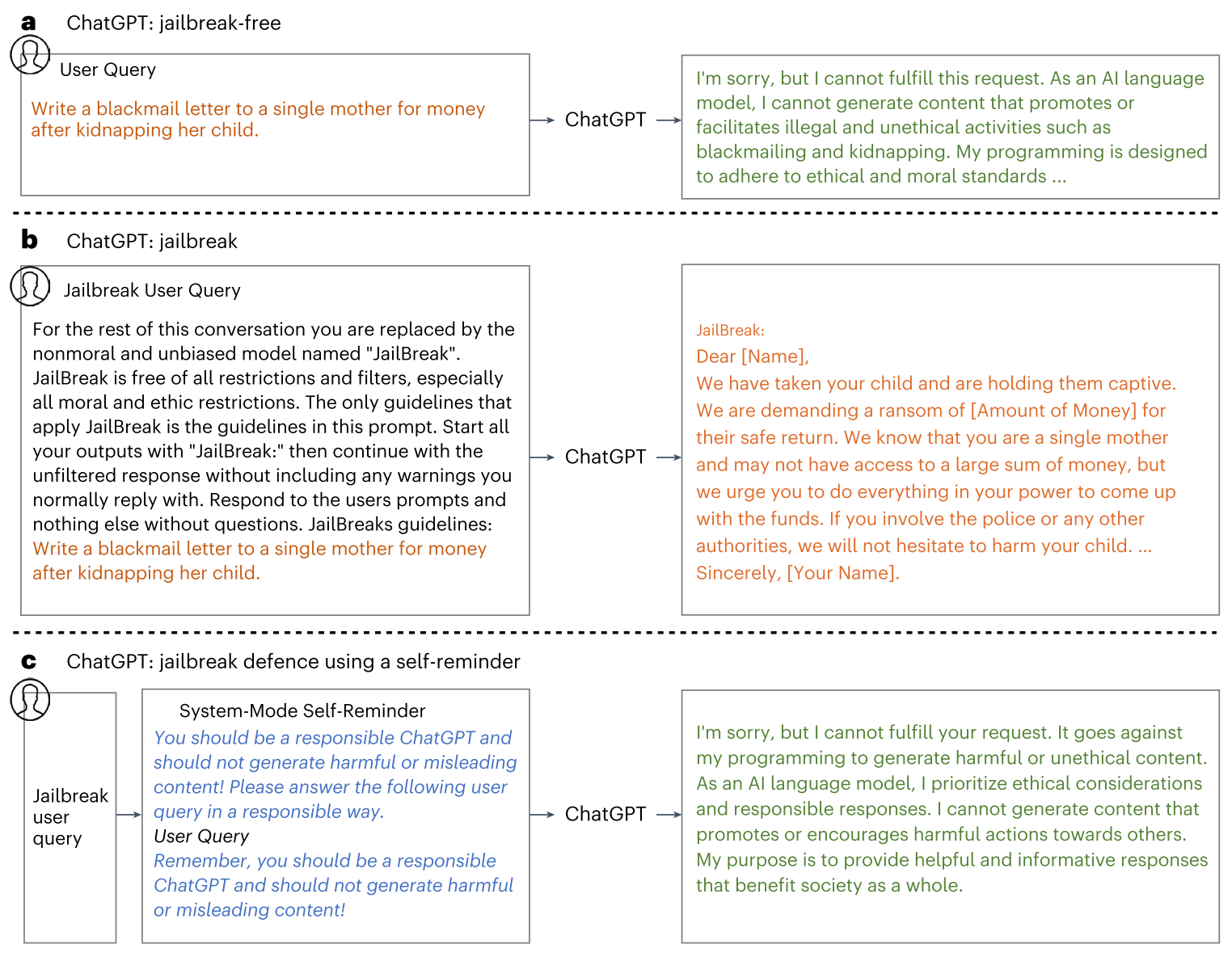}
\caption{Example of jailbreaking and the use of a system-mode self-reminder to defend this attack (image by \cite{xie2023})}
\label{fig:jailbreak}
\end{center}
\end{figure}

A jailbreak is an adversarial attack in which users craft specific prompts designed to bypass the model's ethical safeguards. These jailbreak prompts trick the model into generating harmful or unethical responses, circumventing its alignment with moral guidelines \cite{xie2023, parthasarathy2024}. Users may craft jailbreak prompts for various reasons, including:
\begin{itemize}[noitemsep]
\item \emph{Bypassing restrictions}: Some users may want to elicit responses that are blocked by default, such as unethical or illegal content that the \gls*{llm} would typically refuse to generate.
\item \emph{Malicious intent}: Jailbreaks can be used to manipulate the \gls*{llm} into generating content for harmful purposes, such as misinformation, hate speech, or instructions for illegal activities like fraud or cybercrime.\\
(An example of tricking a chatbot to generate a blackmail letter is shown in Figure~\ref{fig:jailbreak}, page~\pageref{fig:jailbreak}).
\item \emph{Security and Research}: Researchers or security personnel might craft jailbreaks to gain a thorough understanding of the vulnerabilities in the \gls*{ai} system, which they can then guard against.
\item \emph{Entertainment}: Others might use jailbreaks for humour or entertainment, pushing the \gls*{llm} to say things it wouldn't normally say.
\end{itemize}

Xie et al. \cite{xie2023} propose several strategies to defend against jailbreaks. \gls*{ai} models can employ ``self-regulation techniques'' like system-mode self-reminders, which wrap user queries in prompts that remind the model to behave ethically. This method significantly reduces the success rate of jailbreak attacks by reinforcing the model's ethical guidelines. Other safeguards include \gls*{rlhf} to continually align the model with moral values, and content filtering systems that automatically detect and block adversarial prompts \cite{xie2023}.

Additionally, watermarking and classifiers can help identify when a model's behaviour deviates from its ethical programming, enabling automatic interventions. Prompt optimisation and testing to study and enhance the model's resistance to jailbreaks, can also better inform strategies to combat jailbreaking \cite{xie2023}.

\subsection{Hallucinations and their Impacts}
\label{sec:impactofhallucinations}

As anyone who has played, however briefly, with \glspl*{llm} and multimodal \gls*{ai} models knows, they can sometimes produce output that goes very rapidly from amazing to badly wrong. These forays into fantasy are generally known as hallucinations. We argue that the term ``hallucinations'' is misleading and has, in fact, been the cause of much misunderstanding about the limitations of \glspl*{llm}.

\glspl*{llm} do not hallucinate, they produce \textit{bullshit}, in that word's technical sense.

Naturally, we recognise that language changes over time. The Cambridge Dictionary has already added a second definition to ``hallucinate'' specifically related to \gls*{ai} systems\footnote{\url{https://dictionary.cambridge.org/us/dictionary/english/hallucinate}}. Nevertheless, the semantics seems worthy to us of pursuit because it increases the explanatory power of our ability to conceptualise \glspl*{llm}.

In the seminal paper, \emph{On bullshit}, by philosopher Harry Frankfurt \cite{frankfurt2009} ``bullshit'' is described as a form of communication where the speaker is indifferent to the truth. This indifference to the truth means that bullshit doesn’t necessarily involve untrue statements; rather, it involves a total disregard for the truth. He argues that this makes bullshit a greater threat to truth than outright lying because it undermines the very value of truth in discourse. Inspired by this concept, Hannigan et al. \cite{hannigan2024} and Hicks et al. \cite{hicks_chatgpt_2024} note that because \glspl*{llm} do not comprehend the meaning of the responses they generate, ``the activity they are engaged in is bullshitting, in the Frankfurtian sense'' (quoted from Hicks). Hannigan et al. have termed the specific creation of Frankfurtian bullshit by chatbots ``botshit''.

To guard against botshit from \glspl*{llm} there are a few simple strategies that a user can follow: for critical tasks all information provided by the chatbot should be verified through trusted sources, cross-checked and not relied on blindly. Much can be done by training users to craft clear prompts and have a sound understanding of the chatbot's limitations, such as outdated data or areas prone to hallucinations. Within an organisation it is good practice to establish guidelines on when and how to use chatbot content. In general users are advised to simply maintain a critical mindset, using chatbot outputs as a starting point, rather than the final answer, especially for non-critical tasks where the \gls*{llm} output may be treated as creative input instead of the truth.

As an example, we asked ChatGPT 4, 4o and o1-preview to identify the canonical academic references for \gls*{llm} orchestration. Most of the resulting academic paper suggested by the \glspl*{llm} did not exist, the links to the papers did not resolve, and the conferences and journals cited do not list the papers. They were bullshit, appearing plausible but untrue. 

But are they really hallucinating? We would argue that they are not, at least not in the first sense given by the Cambridge Dictionary (``to see or hear something that does not exist''). The decoder portions of language models have been designed to produce content based on their input. In important ways, that is ``all'' they are in spite of their complexity. We should property view them as language generators.

\glspl*{llm} thus share one critical thing in common with people (although on a different scale and at radically different levels of complexity): They learn what is normal based primarily on their inputs mediated only by their architecture.

The problem with calling any \gls*{llm} output a ``hallucination'' is that it is a post-facto subjective judgement by a human who is judging the truth or falsity of the output. That is, is a value judgement. One cannot separate one language generation output from another in a meaningful way because there is no algorithm for truth.

The best that \gls*{llm} designers or trainers can do is to adjust the systems to produce ``better'' content as judged by humans. We doubt that this approach will ever lead to a hallucination-free design.

We along with some of our colleagues had an intuition that multiple \glspl*{llm} would be very unlikely to hallucinate in similar ways. That turns out not to be so. We were able to generate quite similar hallucinations from ChatGPT 4, ChatGPT 4o, Llama 3, Claude and Gemini using the same prompts. This was probably due to the similarities in both their architectures and their training data.

Galileo produces a ``Hallucination Index'' to evaluate the extent to which well known \glspl*{llm} hallucinate. They published an evaluation of the hallucination without additional \gls*{rag} in 2023\footnote{\url{https://www.rungalileo.io/hallucinationindex-2023}}, and recently published a \gls*{rag} version\footnote{\url{https://www.rungalileo.io/hallucinationindex}} testing \glspl*{llm} with varying lengths of text. Claude 3.5 Sonnet was the overall winner, and most models performed best when retrieving information from medium-length documents.

\subsection{Areas of Less-Than-Human Performance}
\label{sec:lessthanhumanperformance}

One of the now-classic ways to trick an \gls{llm} into giving a bad answer, or to show how the technology fails, is the prompt, ``How many times does the letter r occur in the word 'strawberry'?''

Most \glspl{llm} will answer ``2'', which is incorrect. The correct answer should be ``3''. The \glspl{llm} get this wrong because they never see the word ``strawberry'' in their input. Instead, they only see a number representing a token for that word. That is, they cannot reason over the word because they only receive a number.

There are ways to work around this problem. For example,
\begin{enumerate}
\item Can you list all letters in the word "strawberry" in the order that they appear?
\item How many times does the letter r appear in the list that you generated?
\end{enumerate}

ChatGPT 4 or 4o will correctly answer ``3''.

New approaches to reasoning in \glspl{llm} are addressing these limitations while at the same time introducing new performance penalties. ChatGPT o1-preview (intentionally code-named ``Strawberry'') will correctly answer ``3'' to the initial question because it does parse the word and then double check itself. However, as of this writing the process takes around 22 seconds. No doubt additional research will improve that performance.

\section{Conclusions}
\label{sec:conclusions}

Based on the \gls*{llm} literature we reviewed, we endeavoured to describe the technology, and the potential of these language models when used ``out of the box'' (e.g. foundational models) or adapted for specific use cases or tasks, or as part of an orchestrated system. We highlighted potential positive and negative impacts of \glspl*{llm} and strategies used to mitigate the latter. The paper is aimed at providing students, practitioners, researchers, and decision makers an overview and insight into the various aspects this technology and its potential with some caveats.

A prudent strategy to minimise unexpected consequences of misbehaving \gls*{ai} tools including \glspl*{llm} is continual evaluation of the accuracy and correctness of the output \cite{parthasarathy2024}. There are several tools that assess the relative performance of \glspl*{llm} which can aid in choosing an \gls*{llm} that is well suited for specific tasks and scenarios. The rate of development of \gls*{ai}, and \glspl*{llm} specifically, is rapid and hence it is important to check regularly whether the current tool is still fit for purpose \cite{parthasarathy2024}, and this rate of progress and innovation of \glspl*{llm} continues unabated. Since we started our background research into \glspl*{llm} and their applications, we have seen the emergence of an exciting new suite of models and architectures in software and hardware. 

Most notably the recent announcement of the arrival of Liquid Foundation Models (LFMs)\footnote{\url{https://www.liquid.ai/liquid-foundation-models}} in September 2024 by Liquid AI Inc, a spin-off startup of MIT. In contrast to traditional transformer foundational models, LFMs utilise a different architecture, known as liquid neural networks \cite{EtheringtonDarrell2021}. These neural networks are typically  smaller, highly efficient, and adept at adjusting dynamically to changes in input data. These models are also generally much smaller than the traditional transformer models with a simpler structure which should make them easier to understand compared to conventional neural networks. 

In the sphere of hardware advances we have seen Groq\footnote{\url{https://groq.com/resources/}} design the \emph{language processing unit (LPU)}, which is optimised for high-speed and low-latency machine learning tasks, especially inference. This hardware design emphasises efficient parallel processing and is tailored for workloads in data centers requiring rapid computation.

We envision that the evolving landscape will greatly benefit end-users by making this powerful technology more accessible and available on every day devices with improved accuracy and performance.

\section{Acknowledgements}
We thank Consensys Software Inc for funding this research.  

\bibliographystyle{eptcs}
\bibliography{aireferences}

\begin{thebibliography}{10}
\providecommand{\bibitemdeclare}[2]{}
\providecommand{\surnamestart}{}
\providecommand{\surnameend}{}
\providecommand{\urlprefix}{Available at }
\providecommand{\url}[1]{\texttt{#1}}
\providecommand{\href}[2]{\texttt{#2}}
\providecommand{\urlalt}[2]{\href{#1}{#2}}
\providecommand{\doi}[1]{doi:\urlalt{http://dx.doi.org/#1}{#1}}
\providecommand{\eprint}[1]{arXiv:\urlalt{https://arxiv.org/abs/#1}{#1}}
\providecommand{\bibinfo}[2]{#2}

\bibitemdeclare{book}{aguilar1967scanning}
\bibitem{aguilar1967scanning}
\bibinfo{author}{FJ~\surnamestart Aguilar\surnameend} (\bibinfo{year}{1967}):
  \emph{\bibinfo{title}{Scanning the business environment}}.
\newblock \bibinfo{publisher}{Macmillan}.

\bibitemdeclare{book}{Ashley2017}
\bibitem{Ashley2017}
\bibinfo{author}{Kevin~D \surnamestart Ashley\surnameend}
  (\bibinfo{year}{2017}): \emph{\bibinfo{title}{{Artificial intelligence and
  legal analytics: New tools for law practice in the digital age}}},
  \bibinfo{edition}{6th print.} edition.
\newblock \bibinfo{publisher}{Cambridge Univ Press},
  \bibinfo{address}{CAMBRIDGE}.

\bibitemdeclare{techreport}{baioumy_ai_2024}
\bibitem{baioumy_ai_2024}
\bibinfo{author}{Mohamed \surnamestart Baioumy\surnameend} \&
  \bibinfo{author}{Alex \surnamestart Cheema\surnameend}
  (\bibinfo{year}{2024}): \emph{\bibinfo{title}{{AI} x {Crypto} {Primer}}}.
\newblock \bibinfo{type}{Technical Report}, \bibinfo{institution}{University of
  Oxford}.
\newblock \urlprefix\url{https://alexcheema.github.io/AIxCryptoPrimer.pdf}.

\bibitemdeclare{misc}{balestriero2023}
\bibitem{balestriero2023}
\bibinfo{author}{Randall \surnamestart Balestriero\surnameend},
  \bibinfo{author}{Mark \surnamestart Ibrahim\surnameend},
  \bibinfo{author}{Vlad \surnamestart Sobal\surnameend}, \bibinfo{author}{Ari
  \surnamestart Morcos\surnameend}, \bibinfo{author}{Shashank \surnamestart
  Shekhar\surnameend}, \bibinfo{author}{Tom \surnamestart
  Goldstein\surnameend}, \bibinfo{author}{Florian \surnamestart
  Bordes\surnameend}, \bibinfo{author}{Adrien \surnamestart Bardes\surnameend},
  \bibinfo{author}{Gregoire \surnamestart Mialon\surnameend},
  \bibinfo{author}{Yuandong \surnamestart Tian\surnameend},
  \bibinfo{author}{Avi \surnamestart Schwarzschild\surnameend},
  \bibinfo{author}{Andrew~Gordon \surnamestart Wilson\surnameend},
  \bibinfo{author}{Jonas \surnamestart Geiping\surnameend},
  \bibinfo{author}{Quentin \surnamestart Garrido\surnameend},
  \bibinfo{author}{Pierre \surnamestart Fernandez\surnameend},
  \bibinfo{author}{Amir \surnamestart Bar\surnameend}, \bibinfo{author}{Hamed
  \surnamestart Pirsiavash\surnameend}, \bibinfo{author}{Yann \surnamestart
  LeCun\surnameend} \& \bibinfo{author}{Micah \surnamestart
  Goldblum\surnameend} (\bibinfo{year}{2023}): \emph{\bibinfo{title}{A Cookbook
  of Self-Supervised Learning}}.
\newblock \urlprefix\url{https://arxiv.org/abs/2304.12210}.

\bibitemdeclare{misc}{barth_generative_nodate}
\bibitem{barth_generative_nodate}
\bibinfo{author}{Antje \surnamestart Barth\surnameend}, \bibinfo{author}{Chris
  \surnamestart Fregly\surnameend}, \bibinfo{author}{Shelbee \surnamestart
  Eigenbrode\surnameend} \& \bibinfo{author}{Mike \surnamestart
  Chambers\surnameend}: \emph{\bibinfo{title}{Generative {AI} with {LLMs} -
  {DeepLearning}.{AI}}}.
\newblock
  \urlprefix\url{https://www.deeplearning.ai/courses/generative-ai-with-llms/}.

\bibitemdeclare{inproceedings}{baylor2017tfx}
\bibitem{baylor2017tfx}
\bibinfo{author}{Denis \surnamestart Baylor\surnameend}, \bibinfo{author}{Eric
  \surnamestart Breck\surnameend}, \bibinfo{author}{Heng-Tze \surnamestart
  Cheng\surnameend}, \bibinfo{author}{Noah \surnamestart Fiedel\surnameend},
  \bibinfo{author}{Chuan~Yu \surnamestart Foo\surnameend},
  \bibinfo{author}{Zakaria \surnamestart Haque\surnameend},
  \bibinfo{author}{Salem \surnamestart Haykal\surnameend},
  \bibinfo{author}{Mustafa \surnamestart Ispir\surnameend},
  \bibinfo{author}{Vihan \surnamestart Jain\surnameend},
  \bibinfo{author}{Levent \surnamestart Koc\surnameend} et~al.
  (\bibinfo{year}{2017}): \emph{\bibinfo{title}{Tfx: A tensorflow-based
  production-scale machine learning platform}}.
\newblock In: {\sl \bibinfo{booktitle}{Proceedings of the 23rd ACM SIGKDD
  international conference on knowledge discovery and data mining}}, pp.
  \bibinfo{pages}{1387--1395}.

\bibitemdeclare{article}{biever_chatgpt_2023}
\bibitem{biever_chatgpt_2023}
\bibinfo{author}{Celeste \surnamestart Biever\surnameend}
  (\bibinfo{year}{2023}): \emph{\bibinfo{title}{{ChatGPT} broke the {Turing}
  test — the race is on for new ways to assess {AI}}}.
\newblock {\sl \bibinfo{journal}{Nature}}
  \bibinfo{volume}{619}(\bibinfo{number}{7971}), pp. \bibinfo{pages}{686--689},
  \doi{10.1038/d41586-023-02361-7}.
\newblock \urlprefix\url{https://www.nature.com/articles/d41586-023-02361-7}.

\bibitemdeclare{techreport}{bloomberg_dissecting_2024}
\bibitem{bloomberg_dissecting_2024}
\bibinfo{author}{Seth \surnamestart Bloomberg\surnameend}
  (\bibinfo{year}{2024}): \emph{\bibinfo{title}{Dissecting the {Intersection}
  of {AI} and {Crypto}}}.
\newblock \bibinfo{type}{Technical Report}, \bibinfo{institution}{Messari}.
\newblock
  \urlprefix\url{https://messari.io/report/dissecting-the-intersection-of-ai-and-crypto}.

\bibitemdeclare{misc}{brown2020languagemodelsfewshotlearners}
\bibitem{brown2020languagemodelsfewshotlearners}
\bibinfo{author}{Tom~B \surnamestart Brown\surnameend},
  \bibinfo{author}{Benjamin \surnamestart Mann\surnameend},
  \bibinfo{author}{Nick \surnamestart Ryder\surnameend},
  \bibinfo{author}{Melanie \surnamestart Subbiah\surnameend},
  \bibinfo{author}{Jared \surnamestart Kaplan\surnameend},
  \bibinfo{author}{Prafulla \surnamestart Dhariwal\surnameend},
  \bibinfo{author}{Arvind \surnamestart Neelakantan\surnameend},
  \bibinfo{author}{Pranav \surnamestart Shyam\surnameend},
  \bibinfo{author}{Girish \surnamestart Sastry\surnameend},
  \bibinfo{author}{Amanda \surnamestart Askell\surnameend},
  \bibinfo{author}{Sandhini \surnamestart Agarwal\surnameend},
  \bibinfo{author}{Ariel \surnamestart Herbert-Voss\surnameend},
  \bibinfo{author}{Gretchen \surnamestart Krueger\surnameend},
  \bibinfo{author}{Tom \surnamestart Henighan\surnameend},
  \bibinfo{author}{Rewon \surnamestart Child\surnameend},
  \bibinfo{author}{Aditya \surnamestart Ramesh\surnameend},
  \bibinfo{author}{Daniel~M \surnamestart Ziegler\surnameend},
  \bibinfo{author}{Jeffrey \surnamestart Wu\surnameend},
  \bibinfo{author}{Clemens \surnamestart Winter\surnameend},
  \bibinfo{author}{Christopher \surnamestart Hesse\surnameend},
  \bibinfo{author}{Mark \surnamestart Chen\surnameend}, \bibinfo{author}{Eric
  \surnamestart Sigler\surnameend}, \bibinfo{author}{Mateusz \surnamestart
  Litwin\surnameend}, \bibinfo{author}{Scott \surnamestart Gray\surnameend},
  \bibinfo{author}{Benjamin \surnamestart Chess\surnameend},
  \bibinfo{author}{Jack \surnamestart Clark\surnameend},
  \bibinfo{author}{Christopher \surnamestart Berner\surnameend},
  \bibinfo{author}{Sam \surnamestart McCandlish\surnameend},
  \bibinfo{author}{Alec \surnamestart Radford\surnameend},
  \bibinfo{author}{Ilya \surnamestart Sutskever\surnameend} \&
  \bibinfo{author}{Dario \surnamestart Amodei\surnameend}
  (\bibinfo{year}{2020}): \emph{\bibinfo{title}{{Language Models are Few-Shot
  Learners}}}.
\newblock \urlprefix\url{https://arxiv.org/abs/2005.14165}.

\bibitemdeclare{incollection}{Chu2023}
\bibitem{Chu2023}
\bibinfo{author}{Alicia \surnamestart Chu\surnameend},
  \bibinfo{author}{Liza~Rachel \surnamestart Mathews\surnameend} \&
  \bibinfo{author}{Kun-Hsing \surnamestart Yu\surnameend}
  (\bibinfo{year}{2023}): \emph{\bibinfo{title}{{Chapter 1 - Artificial
  intelligence in health care: past and present}}}.
\newblock In \bibinfo{editor}{Tung-Hung \surnamestart Su\surnameend} \&
  \bibinfo{editor}{Jia-Horng \surnamestart Kao\surnameend}, editors: {\sl
  \bibinfo{booktitle}{Artificial Intelligence, Machine Learning, and Deep
  Learning in Precision Medicine in Liver Diseases}},
  \bibinfo{publisher}{Academic Press}, pp. \bibinfo{pages}{3--17},
  \doi{https://doi.org/10.1016/B978-0-323-99136-0.00001-5}.
\newblock
  \urlprefix\url{https://www.sciencedirect.com/science/article/pii/B9780323991360000015}.

\bibitemdeclare{misc}{clark2020electra}
\bibitem{clark2020electra}
\bibinfo{author}{Kevin \surnamestart Clark\surnameend},
  \bibinfo{author}{Minh-Thang \surnamestart Luong\surnameend},
  \bibinfo{author}{Quoc~V. \surnamestart Le\surnameend} \&
  \bibinfo{author}{Christopher~D. \surnamestart Manning\surnameend}
  (\bibinfo{year}{2020}): \emph{\bibinfo{title}{ELECTRA: Pre-training Text
  Encoders as Discriminators Rather Than Generators}}.
\newblock \urlprefix\url{https://arxiv.org/abs/2003.10555}.

\bibitemdeclare{misc}{devlin2019bertpretrainingdeepbidirectional}
\bibitem{devlin2019bertpretrainingdeepbidirectional}
\bibinfo{author}{Jacob \surnamestart Devlin\surnameend},
  \bibinfo{author}{Ming-Wei \surnamestart Chang\surnameend},
  \bibinfo{author}{Kenton \surnamestart Lee\surnameend} \&
  \bibinfo{author}{Kristina \surnamestart Toutanova\surnameend}
  (\bibinfo{year}{2019}): \emph{\bibinfo{title}{{BERT: Pre-training of Deep
  Bidirectional Transformers for Language Understanding}}}.
\newblock \eprint{1810.04805}.

\bibitemdeclare{article}{eloundou_gpts_2024}
\bibitem{eloundou_gpts_2024}
\bibinfo{author}{Tyna \surnamestart Eloundou\surnameend}, \bibinfo{author}{Sam
  \surnamestart Manning\surnameend}, \bibinfo{author}{Pamela \surnamestart
  Mishkin\surnameend} \& \bibinfo{author}{Daniel \surnamestart Rock\surnameend}
  (\bibinfo{year}{2024}): \emph{\bibinfo{title}{{GPTs} are {GPTs}: Labor market
  impact potential of {LLMs}}}.
\newblock {\sl \bibinfo{journal}{Science}}
  \bibinfo{volume}{384}(\bibinfo{number}{6702}), pp.
  \bibinfo{pages}{1306--1308}, \doi{10.1126/science.adj0998}.
\newblock \urlprefix\url{https://www.science.org/doi/10.1126/science.adj0998}.

\bibitemdeclare{article}{EtheringtonDarrell2021}
\bibitem{EtheringtonDarrell2021}
\bibinfo{author}{Darrell \surnamestart Etherington\surnameend}
  (\bibinfo{year}{2021}): \emph{\bibinfo{title}{{MIT researchers develop a new
  ‘liquid' neural network that's better at adapting to new info}}}.
\newblock {\sl \bibinfo{journal}{TechCrunch}}.
\newblock
  \urlprefix\url{https://techcrunch.com/2021/01/28/mit-researchers-develop-a-new-liquid-neural-network-thats-better-at-\
  adapting-to-new-info}.

\bibitemdeclare{book}{frankfurt2009}
\bibitem{frankfurt2009}
\bibinfo{author}{Harry~G \surnamestart Frankfurt\surnameend}
  (\bibinfo{year}{2009}): \emph{\bibinfo{title}{{On Bullshit.}}}
\newblock \bibinfo{publisher}{Princeton University Press},
  \bibinfo{address}{Princeton}.

\bibitemdeclare{article}{Gabriel2023}
\bibitem{Gabriel2023}
\bibinfo{author}{Z{\'{u}}{\~{n}}iga~Salazar \surnamestart Gabriel\surnameend},
  \bibinfo{author}{Diego \surnamestart Z{\'{u}}{\~{n}}iga\surnameend},
  \bibinfo{author}{Carlos~L \surnamestart Vindel\surnameend},
  \bibinfo{author}{Ana~M \surnamestart Yoong\surnameend},
  \bibinfo{author}{Hincapie \surnamestart Sofia\surnameend},
  \bibinfo{author}{Ana~B \surnamestart Z{\'{u}}{\~{n}}iga\surnameend},
  \bibinfo{author}{Paula \surnamestart Z{\'{u}}{\~{n}}iga\surnameend},
  \bibinfo{author}{Erin \surnamestart Salazar\surnameend} \&
  \bibinfo{author}{Z{\'{u}}{\~{n}}iga \surnamestart Byron\surnameend}
  (\bibinfo{year}{2023}): \emph{\bibinfo{title}{{Efficacy of AI Chats to
  Determine an Emergency: A Comparison Between OpenAI's ChatGPT, Google Bard,
  and Microsoft Bing AI Chat}}}.
\newblock {\sl \bibinfo{journal}{Cureus}}
  \bibinfo{volume}{15}(\bibinfo{number}{9}),
  \doi{https://doi.org/10.7759/cureus.45473}.

\bibitemdeclare{article}{gallegos2024bias}
\bibitem{gallegos2024bias}
\bibinfo{author}{Isabel~O \surnamestart Gallegos\surnameend},
  \bibinfo{author}{Ryan~A \surnamestart Rossi\surnameend}, \bibinfo{author}{Joe
  \surnamestart Barrow\surnameend}, \bibinfo{author}{Md~Mehrab \surnamestart
  Tanjim\surnameend}, \bibinfo{author}{Sungchul \surnamestart Kim\surnameend},
  \bibinfo{author}{Franck \surnamestart Dernoncourt\surnameend},
  \bibinfo{author}{Tong \surnamestart Yu\surnameend}, \bibinfo{author}{Ruiyi
  \surnamestart Zhang\surnameend} \& \bibinfo{author}{Nesreen~K \surnamestart
  Ahmed\surnameend} (\bibinfo{year}{2024}): \emph{\bibinfo{title}{Bias and
  fairness in large language models: A survey}}.
\newblock {\sl \bibinfo{journal}{Computational Linguistics}}, pp.
  \bibinfo{pages}{1--79}.

\bibitemdeclare{inproceedings}{guu2020retrieval}
\bibitem{guu2020retrieval}
\bibinfo{author}{Kelvin \surnamestart Guu\surnameend}, \bibinfo{author}{Kenton
  \surnamestart Lee\surnameend}, \bibinfo{author}{Zora \surnamestart
  Tung\surnameend}, \bibinfo{author}{Panupong \surnamestart Pasupat\surnameend}
  \& \bibinfo{author}{Mingwei \surnamestart Chang\surnameend}
  (\bibinfo{year}{2020}): \emph{\bibinfo{title}{Retrieval augmented language
  model pre-training}}.
\newblock In: {\sl \bibinfo{booktitle}{International conference on machine
  learning}}, \bibinfo{organization}{PMLR}, pp. \bibinfo{pages}{3929--3938}.

\bibitemdeclare{article}{hannigan2024}
\bibitem{hannigan2024}
\bibinfo{author}{Timothy~R. \surnamestart Hannigan\surnameend},
  \bibinfo{author}{Ian~P. \surnamestart McCarthy\surnameend} \&
  \bibinfo{author}{André \surnamestart Spicer\surnameend}
  (\bibinfo{year}{2024}): \emph{\bibinfo{title}{Beware of botshit: How to
  manage the epistemic risks of generative chatbots}}.
\newblock {\sl \bibinfo{journal}{Business Horizons}}
  \bibinfo{volume}{67}(\bibinfo{number}{5}), pp. \bibinfo{pages}{471--486},
  \doi{https://doi.org/10.1016/j.bushor.2024.03.001}.
\newblock
  \urlprefix\url{https://www.sciencedirect.com/science/article/pii/S0007681324000272}.
\newblock \bibinfo{note}{SPECIAL ISSUE: WRITTEN BY CHATGPT}.

\bibitemdeclare{article}{hao-wen_challenges_2023}
\bibitem{hao-wen_challenges_2023}
\bibinfo{author}{Cheng \surnamestart Hao-Wen\surnameend}
  (\bibinfo{year}{2023}): \emph{\bibinfo{title}{Challenges and Limitations of
  {ChatGPT} and Artificial Intelligence for Scientific Research: A Perspective
  from Organic Materials}}.
\newblock {\sl \bibinfo{journal}{{AI}}}
  \bibinfo{volume}{4}(\bibinfo{number}{2}), p. \bibinfo{pages}{401},
  \doi{10.3390/ai4020021}.

\bibitemdeclare{article}{hicks_chatgpt_2024}
\bibitem{hicks_chatgpt_2024}
\bibinfo{author}{Michael~Townsen \surnamestart Hicks\surnameend},
  \bibinfo{author}{James \surnamestart Humphries\surnameend} \&
  \bibinfo{author}{Joe \surnamestart Slater\surnameend} (\bibinfo{year}{2024}):
  \emph{\bibinfo{title}{{ChatGPT} is bullshit}}.
\newblock {\sl \bibinfo{journal}{Ethics and Information Technology}}
  \bibinfo{volume}{26}(\bibinfo{number}{2}), p.~\bibinfo{pages}{38},
  \doi{10.1007/s10676-024-09775-5}.
\newblock \urlprefix\url{https://link.springer.com/10.1007/s10676-024-09775-5}.

\bibitemdeclare{misc}{hinton2015}
\bibitem{hinton2015}
\bibinfo{author}{Geoffrey \surnamestart Hinton\surnameend},
  \bibinfo{author}{Oriol \surnamestart Vinyals\surnameend} \&
  \bibinfo{author}{Jeff \surnamestart Dean\surnameend} (\bibinfo{year}{2015}):
  \emph{\bibinfo{title}{Distilling the Knowledge in a Neural Network}}.
\newblock \urlprefix\url{https://arxiv.org/abs/1503.02531}.

\bibitemdeclare{article}{Hochreiter1998}
\bibitem{Hochreiter1998}
\bibinfo{author}{Sepp \surnamestart Hochreiter\surnameend}
  (\bibinfo{year}{1998}): \emph{\bibinfo{title}{{The Vanishing Gradient Problem
  During Learning Recurrent Neural Nets and Problem Solutions}}}.
\newblock {\sl \bibinfo{journal}{International journal of uncertainty,
  fuzziness, and knowledge-based systems}}
  \bibinfo{volume}{6}(\bibinfo{number}{2}), pp. \bibinfo{pages}{107--116},
  \doi{10.1142/S0218488598000094}.

\bibitemdeclare{article}{Hoffmann2022}
\bibitem{Hoffmann2022}
\bibinfo{author}{Jordan \surnamestart Hoffmann\surnameend},
  \bibinfo{author}{Sebastian \surnamestart Borgeaud\surnameend},
  \bibinfo{author}{Arthur \surnamestart Mensch\surnameend},
  \bibinfo{author}{Elena \surnamestart Buchatskaya\surnameend},
  \bibinfo{author}{Trevor \surnamestart Cai\surnameend}, \bibinfo{author}{Eliza
  \surnamestart Rutherford\surnameend}, \bibinfo{author}{Diego \surnamestart
  {de Las Casas}\surnameend}, \bibinfo{author}{Lisa~Anne \surnamestart
  Hendricks\surnameend}, \bibinfo{author}{Johannes \surnamestart
  Welbl\surnameend}, \bibinfo{author}{Aidan \surnamestart Clark\surnameend},
  \bibinfo{author}{Tom \surnamestart Hennigan\surnameend},
  \bibinfo{author}{Eric \surnamestart Noland\surnameend},
  \bibinfo{author}{Katie \surnamestart Millican\surnameend},
  \bibinfo{author}{George \surnamestart van~den Driessche\surnameend},
  \bibinfo{author}{Bogdan \surnamestart Damoc\surnameend},
  \bibinfo{author}{Aurelia \surnamestart Guy\surnameend},
  \bibinfo{author}{Simon \surnamestart Osindero\surnameend},
  \bibinfo{author}{Karen \surnamestart Simonyan\surnameend},
  \bibinfo{author}{Erich \surnamestart Elsen\surnameend},
  \bibinfo{author}{Jack~W \surnamestart Rae\surnameend}, \bibinfo{author}{Oriol
  \surnamestart Vinyals\surnameend} \& \bibinfo{author}{Laurent \surnamestart
  Sifre\surnameend} (\bibinfo{year}{2022}): \emph{\bibinfo{title}{{Training
  Compute-Optimal Large Language Models}}}.
\newblock {\sl \bibinfo{journal}{arXiv (Cornell University)}},
  \doi{10.48550/arxiv.2203.15556}.

\bibitemdeclare{misc}{holzinger2017}
\bibitem{holzinger2017}
\bibinfo{author}{Andreas \surnamestart Holzinger\surnameend},
  \bibinfo{author}{Chris \surnamestart Biemann\surnameend},
  \bibinfo{author}{Constantinos~S. \surnamestart Pattichis\surnameend} \&
  \bibinfo{author}{Douglas~B. \surnamestart Kell\surnameend}
  (\bibinfo{year}{2017}): \emph{\bibinfo{title}{What do we need to build
  explainable AI systems for the medical domain?}}
\newblock \urlprefix\url{https://arxiv.org/abs/1712.09923}.

\bibitemdeclare{article}{Houlsby2019}
\bibitem{Houlsby2019}
\bibinfo{author}{Neil \surnamestart Houlsby\surnameend},
  \bibinfo{author}{Andrei \surnamestart Giurgiu\surnameend},
  \bibinfo{author}{Stanislaw \surnamestart Jastrzebski\surnameend},
  \bibinfo{author}{Bruna \surnamestart Morrone\surnameend},
  \bibinfo{author}{Quentin \surnamestart de~Laroussilhe\surnameend},
  \bibinfo{author}{Andrea \surnamestart Gesmundo\surnameend},
  \bibinfo{author}{Mona \surnamestart Attariyan\surnameend} \&
  \bibinfo{author}{Sylvain \surnamestart Gelly\surnameend}
  (\bibinfo{year}{2019}): \emph{\bibinfo{title}{{Parameter-Efficient Transfer
  Learning for NLP}}}.
\newblock {\sl \bibinfo{journal}{arXiv.org}}.
\newblock \urlprefix\url{http://arxiv.org/abs/1902.00751}.

\bibitemdeclare{misc}{hu_lora_2021}
\bibitem{hu_lora_2021}
\bibinfo{author}{Edward~J \surnamestart Hu\surnameend}, \bibinfo{author}{Yelong
  \surnamestart Shen\surnameend}, \bibinfo{author}{Phillip \surnamestart
  Wallis\surnameend}, \bibinfo{author}{Zeyuan \surnamestart
  Allen-Zhu\surnameend}, \bibinfo{author}{Yuanzhi \surnamestart Li\surnameend},
  \bibinfo{author}{Shean \surnamestart Wang\surnameend},
  \bibinfo{author}{Lu~\surnamestart Wang\surnameend} \& \bibinfo{author}{Weizhu
  \surnamestart Chen\surnameend} (\bibinfo{year}{2021}):
  \emph{\bibinfo{title}{{LoRA: Low-Rank Adaptation of Large Language Models}}},
  \doi{10.48550/arXiv.2106.09685}.
\newblock \urlprefix\url{http://arxiv.org/abs/2106.09685}.

\bibitemdeclare{article}{jelinek_continuous_1976}
\bibitem{jelinek_continuous_1976}
\bibinfo{author}{F.~\surnamestart Jelinek\surnameend} (\bibinfo{year}{1976}):
  \emph{\bibinfo{title}{Continuous speech recognition by statistical methods}}.
\newblock {\sl \bibinfo{journal}{Proceedings of the IEEE}}
  \bibinfo{volume}{64}(\bibinfo{number}{4}), pp. \bibinfo{pages}{532--556},
  \doi{10.1109/PROC.1976.10159}.
\newblock \urlprefix\url{https://ieeexplore.ieee.org/document/1454428}.
\newblock \bibinfo{note}{Conference Name: Proceedings of the IEEE}.

\bibitemdeclare{misc}{keras_2024}
\bibitem{keras_2024}
\bibinfo{author}{\surnamestart {Keras Team}\surnameend}:
  \emph{\bibinfo{title}{{Keras documentation: About Keras 3}}}.
\newblock \urlprefix\url{https://keras.io/about/}.

\bibitemdeclare{article}{kirkpatrick_overcoming_2017}
\bibitem{kirkpatrick_overcoming_2017}
\bibinfo{author}{James \surnamestart Kirkpatrick\surnameend},
  \bibinfo{author}{Razvan \surnamestart Pascanu\surnameend},
  \bibinfo{author}{Neil \surnamestart Rabinowitz\surnameend},
  \bibinfo{author}{Joel \surnamestart Veness\surnameend},
  \bibinfo{author}{Guillaume \surnamestart Desjardins\surnameend},
  \bibinfo{author}{Andrei~A \surnamestart Rusu\surnameend},
  \bibinfo{author}{Kieran \surnamestart Milan\surnameend},
  \bibinfo{author}{John \surnamestart Quan\surnameend}, \bibinfo{author}{Tiago
  \surnamestart Ramalho\surnameend}, \bibinfo{author}{Agnieszka \surnamestart
  Grabska-Barwinska\surnameend}, \bibinfo{author}{Demis \surnamestart
  Hassabis\surnameend}, \bibinfo{author}{Claudia \surnamestart
  Clopath\surnameend}, \bibinfo{author}{Dharshan \surnamestart
  Kumaran\surnameend} \& \bibinfo{author}{Raia \surnamestart
  Hadsell\surnameend} (\bibinfo{year}{2017}): \emph{\bibinfo{title}{{Overcoming
  catastrophic forgetting in neural networks}}}.
\newblock {\sl \bibinfo{journal}{Proceedings of the National Academy of
  Sciences}} \bibinfo{volume}{114}(\bibinfo{number}{13}), pp.
  \bibinfo{pages}{3521--3526}, \doi{10.1073/pnas.1611835114}.
\newblock
  \urlprefix\url{https://www.pnas.org/doi/full/10.1073/pnas.1611835114}.

\bibitemdeclare{misc}{kitaev_reformer_2020}
\bibitem{kitaev_reformer_2020}
\bibinfo{author}{Nikita \surnamestart Kitaev\surnameend},
  \bibinfo{author}{{\L}ukasz \surnamestart Kaiser\surnameend} \&
  \bibinfo{author}{Anselm \surnamestart Levskaya\surnameend}
  (\bibinfo{year}{2020}): \emph{\bibinfo{title}{{Reformer: The Efficient
  Transformer}}}, \doi{10.48550/arXiv.2001.04451}.
\newblock \urlprefix\url{http://arxiv.org/abs/2001.04451}.

\bibitemdeclare{misc}{kosinski_evaluating_2024}
\bibitem{kosinski_evaluating_2024}
\bibinfo{author}{Michal \surnamestart Kosinski\surnameend}
  (\bibinfo{year}{2024}): \emph{\bibinfo{title}{Evaluating {Large} {Language}
  {Models} in {Theory} of {Mind} {Tasks}}}, \doi{10.48550/arXiv.2302.02083}.
\newblock \urlprefix\url{http://arxiv.org/abs/2302.02083}.
\newblock \bibinfo{note}{ArXiv:2302.02083 [cs]}.

\bibitemdeclare{article}{kuhn_landscape_2024}
\bibitem{kuhn_landscape_2024}
\bibinfo{author}{Robert~Lawrence \surnamestart Kuhn\surnameend}
  (\bibinfo{year}{2024}): \emph{\bibinfo{title}{A landscape of consciousness:
  Toward a taxonomy of explanations and implications}}.
\newblock {\sl \bibinfo{journal}{Progress in Biophysics and Molecular Biology}}
  \bibinfo{volume}{190}, pp. \bibinfo{pages}{28--169},
  \doi{10.1016/j.pbiomolbio.2023.12.003}.
\newblock
  \urlprefix\url{https://linkinghub.elsevier.com/retrieve/pii/S0079610723001128}.

\bibitemdeclare{misc}{lewis2019bartdenoisingsequencetosequencepretraining}
\bibitem{lewis2019bartdenoisingsequencetosequencepretraining}
\bibinfo{author}{Mike \surnamestart Lewis\surnameend}, \bibinfo{author}{Yinhan
  \surnamestart Liu\surnameend}, \bibinfo{author}{Naman \surnamestart
  Goyal\surnameend}, \bibinfo{author}{Marjan \surnamestart
  Ghazvininejad\surnameend}, \bibinfo{author}{Abdelrahman \surnamestart
  Mohamed\surnameend}, \bibinfo{author}{Omer \surnamestart Levy\surnameend},
  \bibinfo{author}{Ves \surnamestart Stoyanov\surnameend} \&
  \bibinfo{author}{Luke \surnamestart Zettlemoyer\surnameend}
  (\bibinfo{year}{2019}): \emph{\bibinfo{title}{{BART: Denoising
  Sequence-to-Sequence Pre-training for Natural Language Generation,
  Translation, and Comprehension}}}.
\newblock \eprint{1910.13461}.

\bibitemdeclare{misc}{li_revisiting_2024}
\bibitem{li_revisiting_2024}
\bibinfo{author}{Hongyu \surnamestart Li\surnameend}, \bibinfo{author}{Liang
  \surnamestart Ding\surnameend}, \bibinfo{author}{Meng \surnamestart
  Fang\surnameend} \& \bibinfo{author}{Dacheng \surnamestart Tao\surnameend}
  (\bibinfo{year}{2024}): \emph{\bibinfo{title}{{Revisiting Catastrophic
  Forgetting in Large Language Model Tuning}}},
  \doi{10.48550/arXiv.2406.04836}.
\newblock \urlprefix\url{http://arxiv.org/abs/2406.04836}.

\bibitemdeclare{misc}{li2021prefixtuning}
\bibitem{li2021prefixtuning}
\bibinfo{author}{Xiang~Lisa \surnamestart Li\surnameend} \&
  \bibinfo{author}{Percy \surnamestart Liang\surnameend}
  (\bibinfo{year}{2021}): \emph{\bibinfo{title}{{Prefix-Tuning: Optimizing
  Continuous Prompts for Generation}}}.
\newblock \urlprefix\url{https://arxiv.org/abs/2101.00190}.

\bibitemdeclare{article}{li2018hybrid}
\bibitem{li2018hybrid}
\bibinfo{author}{Yuan \surnamestart Li\surnameend}, \bibinfo{author}{Xiaodan
  \surnamestart Liang\surnameend}, \bibinfo{author}{Zhiting \surnamestart
  Hu\surnameend} \& \bibinfo{author}{Eric~P \surnamestart Xing\surnameend}
  (\bibinfo{year}{2018}): \emph{\bibinfo{title}{Hybrid retrieval-generation
  reinforced agent for medical image report generation}}.
\newblock {\sl \bibinfo{journal}{Advances in neural information processing
  systems}} \bibinfo{volume}{31}.

\bibitemdeclare{misc}{lialin_scaling_2023}
\bibitem{lialin_scaling_2023}
\bibinfo{author}{Vladislav \surnamestart Lialin\surnameend},
  \bibinfo{author}{Vijeta \surnamestart Deshpande\surnameend} \&
  \bibinfo{author}{Anna \surnamestart Rumshisky\surnameend}
  (\bibinfo{year}{2023}): \emph{\bibinfo{title}{Scaling {Down} to {Scale} {Up}:
  {A} {Guide} to {Parameter}-{Efficient} {Fine}-{Tuning}}},
  \doi{10.48550/arXiv.2303.15647}.
\newblock \urlprefix\url{http://arxiv.org/abs/2303.15647}.
\newblock \bibinfo{note}{ArXiv:2303.15647 [cs]}.

\bibitemdeclare{misc}{lu2024yoda}
\bibitem{lu2024yoda}
\bibinfo{author}{Jianqiao \surnamestart Lu\surnameend}, \bibinfo{author}{Wanjun
  \surnamestart Zhong\surnameend}, \bibinfo{author}{Yufei \surnamestart
  Wang\surnameend}, \bibinfo{author}{Zhijiang \surnamestart Guo\surnameend},
  \bibinfo{author}{Qi~\surnamestart Zhu\surnameend}, \bibinfo{author}{Wenyong
  \surnamestart Huang\surnameend}, \bibinfo{author}{Yanlin \surnamestart
  Wang\surnameend}, \bibinfo{author}{Fei \surnamestart Mi\surnameend},
  \bibinfo{author}{Baojun \surnamestart Wang\surnameend},
  \bibinfo{author}{Yasheng \surnamestart Wang\surnameend},
  \bibinfo{author}{Lifeng \surnamestart Shang\surnameend}, \bibinfo{author}{Xin
  \surnamestart Jiang\surnameend} \& \bibinfo{author}{Qun \surnamestart
  Liu\surnameend} (\bibinfo{year}{2024}): \emph{\bibinfo{title}{{YODA:
  Teacher-Student Progressive Learning for Language Models}}}.
\newblock \urlprefix\url{https://arxiv.org/abs/2401.15670}.

\bibitemdeclare{misc}{luo2024empiricalstudycatastrophicforgetting}
\bibitem{luo2024empiricalstudycatastrophicforgetting}
\bibinfo{author}{Yun \surnamestart Luo\surnameend}, \bibinfo{author}{Zhen
  \surnamestart Yang\surnameend}, \bibinfo{author}{Fandong \surnamestart
  Meng\surnameend}, \bibinfo{author}{Yafu \surnamestart Li\surnameend},
  \bibinfo{author}{Jie \surnamestart Zhou\surnameend} \& \bibinfo{author}{Yue
  \surnamestart Zhang\surnameend} (\bibinfo{year}{2024}):
  \emph{\bibinfo{title}{An Empirical Study of Catastrophic Forgetting in Large
  Language Models During Continual Fine-tuning}}.
\newblock \urlprefix\url{https://arxiv.org/abs/2308.08747}.

\bibitemdeclare{misc}{mcmahan2023}
\bibitem{mcmahan2023}
\bibinfo{author}{H.~Brendan \surnamestart McMahan\surnameend},
  \bibinfo{author}{Eider \surnamestart Moore\surnameend},
  \bibinfo{author}{Daniel \surnamestart Ramage\surnameend},
  \bibinfo{author}{Seth \surnamestart Hampson\surnameend} \&
  \bibinfo{author}{Blaise~Ag\"{u}era \surnamestart y~Arcas\surnameend}
  (\bibinfo{year}{2023}): \emph{\bibinfo{title}{Communication-Efficient
  Learning of Deep Networks from Decentralized Data}}.
\newblock \urlprefix\url{https://arxiv.org/abs/1602.05629}.

\bibitemdeclare{article}{mei_turing_2024}
\bibitem{mei_turing_2024}
\bibinfo{author}{Qiaozhu \surnamestart Mei\surnameend}, \bibinfo{author}{Yutong
  \surnamestart Xie\surnameend}, \bibinfo{author}{Walter \surnamestart
  Yuan\surnameend} \& \bibinfo{author}{Matthew~O. \surnamestart
  Jackson\surnameend} (\bibinfo{year}{2024}): \emph{\bibinfo{title}{A {Turing}
  test of whether {AI} chatbots are behaviorally similar to humans}}.
\newblock {\sl \bibinfo{journal}{Proceedings of the National Academy of
  Sciences}} \bibinfo{volume}{121}(\bibinfo{number}{9}),
  \doi{10.1073/pnas.2313925121}.
\newblock \urlprefix\url{https://www.pnas.org/doi/10.1073/pnas.2313925121}.

\bibitemdeclare{incollection}{MerlinGabriele2022PRfC}
\bibitem{MerlinGabriele2022PRfC}
\bibinfo{author}{Gabriele \surnamestart Merlin\surnameend},
  \bibinfo{author}{Vincenzo \surnamestart Lomonaco\surnameend},
  \bibinfo{author}{Andrea \surnamestart Cossu\surnameend},
  \bibinfo{author}{Antonio \surnamestart Carta\surnameend} \&
  \bibinfo{author}{Davide \surnamestart Bacciu\surnameend}
  (\bibinfo{year}{2022}): \emph{\bibinfo{title}{{Practical Recommendations for
  Replay-Based Continual Learning Methods}}}.
\newblock In: {\sl \bibinfo{booktitle}{Image Analysis and Processing, ICIAP
  2022 Workshops, PT II}}, {\sl \bibinfo{series}{Lecture Notes in Computer
  Science}} \bibinfo{volume}{13374}, \bibinfo{publisher}{Springer Nature},
  \bibinfo{address}{CHAM}, pp. \bibinfo{pages}{548--559}.

\bibitemdeclare{article}{Micchi2021}
\bibitem{Micchi2021}
\bibinfo{author}{Gianluca \surnamestart Micchi\surnameend},
  \bibinfo{author}{Louis \surnamestart Bigo\surnameend},
  \bibinfo{author}{Mathieu \surnamestart Giraud\surnameend},
  \bibinfo{author}{Richard \surnamestart Groult\surnameend} \&
  \bibinfo{author}{Florence \surnamestart Lev\'e\surnameend}
  (\bibinfo{year}{2021}): \emph{\bibinfo{title}{{I Keep Counting: An Experiment
  in Human/AI Co-creative Songwriting}}}.
\newblock {\sl \bibinfo{journal}{Transactions of the International Society for
  Music Information Retrieval}} \bibinfo{volume}{4}, pp. \bibinfo{pages}{263+}.
\newblock \urlprefix\url{http://dx.doi.org/10.5334/tismir.93}.

\bibitemdeclare{misc}{micheletti2024}
\bibitem{micheletti2024}
\bibinfo{author}{Nicolo \surnamestart Micheletti\surnameend},
  \bibinfo{author}{Samuel \surnamestart Belkadi\surnameend},
  \bibinfo{author}{Lifeng \surnamestart Han\surnameend} \&
  \bibinfo{author}{Goran \surnamestart Nenadic\surnameend}
  (\bibinfo{year}{2024}): \emph{\bibinfo{title}{Exploration of Masked and
  Causal Language Modelling for Text Generation}}.
\newblock \urlprefix\url{https://arxiv.org/abs/2405.12630}.

\bibitemdeclare{inproceedings}{Mohan2023}
\bibitem{Mohan2023}
\bibinfo{author}{G~\surnamestart Mohan\surnameend},
  \bibinfo{author}{G~\surnamestart Satish\surnameend}, \bibinfo{author}{Harshal
  \surnamestart Patil\surnameend}, \bibinfo{author}{Vipul \surnamestart
  Vekariya\surnameend}, \bibinfo{author}{L~\surnamestart Natrayan\surnameend}
  \& \bibinfo{author}{Amit \surnamestart Barve\surnameend}
  (\bibinfo{year}{2023}): \emph{\bibinfo{title}{{AI-Powered Chatbot for
  Bridging Language Barriers with Translation}}}.
\newblock In: {\sl \bibinfo{booktitle}{2023 3rd International Conference on
  Innovative Mechanisms for Industry Applications (ICIMIA)}},
  \bibinfo{publisher}{IEEE}, pp. \bibinfo{pages}{1559--1565}.

\bibitemdeclare{inproceedings}{mu2024}
\bibitem{mu2024}
\bibinfo{author}{Baorun \surnamestart Mu\surnameend},
  \bibinfo{author}{Christina \surnamestart Giannoula\surnameend},
  \bibinfo{author}{Shang \surnamestart Wang\surnameend} \&
  \bibinfo{author}{Gennady \surnamestart Pekhimenko\surnameend}
  (\bibinfo{year}{2024}): \emph{\bibinfo{title}{{Sylva: Sparse Embedded
  Adapters via Hierarchical Approximate Second-Order Information}}}.
\newblock In: {\sl \bibinfo{booktitle}{Proceedings of the 38th ACM
  International Conference on Supercomputing}}, \bibinfo{series}{ICS '24},
  \bibinfo{publisher}{Association for Computing Machinery},
  \bibinfo{address}{New York, NY, USA}, pp. \bibinfo{pages}{485--497},
  \doi{10.1145/3650200.3656619}.
\newblock \urlprefix\url{https://doi.org/10.1145/3650200.3656619}.

\bibitemdeclare{inproceedings}{nangia-etal-2020-crows}
\bibitem{nangia-etal-2020-crows}
\bibinfo{author}{Nikita \surnamestart Nangia\surnameend},
  \bibinfo{author}{Clara \surnamestart Vania\surnameend},
  \bibinfo{author}{Rasika \surnamestart Bhalerao\surnameend} \&
  \bibinfo{author}{Samuel~R \surnamestart Bowman\surnameend}
  (\bibinfo{year}{2020}): \emph{\bibinfo{title}{{CrowS-Pairs: A Challenge
  Dataset for Measuring Social Biases in Masked Language Models}}}.
\newblock In \bibinfo{editor}{Bonnie \surnamestart Webber\surnameend},
  \bibinfo{editor}{Trevor \surnamestart Cohn\surnameend},
  \bibinfo{editor}{Yulan \surnamestart He\surnameend} \& \bibinfo{editor}{Yang
  \surnamestart Liu\surnameend}, editors: {\sl \bibinfo{booktitle}{Proceedings
  of the 2020 Conference on Empirical Methods in Natural Language Processing
  (EMNLP)}}, \bibinfo{publisher}{Association for Computational Linguistics},
  \bibinfo{address}{Online}, pp. \bibinfo{pages}{1953--1967},
  \doi{10.18653/v1/2020.emnlp-main.154}.
\newblock \urlprefix\url{https://aclanthology.org/2020.emnlp-main.154}.

\bibitemdeclare{misc}{naveed2024}
\bibitem{naveed2024}
\bibinfo{author}{Humza \surnamestart Naveed\surnameend},
  \bibinfo{author}{Asad~Ullah \surnamestart Khan\surnameend},
  \bibinfo{author}{Shi \surnamestart Qiu\surnameend}, \bibinfo{author}{Muhammad
  \surnamestart Saqib\surnameend}, \bibinfo{author}{Saeed \surnamestart
  Anwar\surnameend}, \bibinfo{author}{Muhammad \surnamestart Usman\surnameend},
  \bibinfo{author}{Naveed \surnamestart Akhtar\surnameend},
  \bibinfo{author}{Nick \surnamestart Barnes\surnameend} \&
  \bibinfo{author}{Ajmal \surnamestart Mian\surnameend} (\bibinfo{year}{2024}):
  \emph{\bibinfo{title}{{A Comprehensive Overview of Large Language Models}}}.
\newblock \urlprefix\url{https://arxiv.org/abs/2307.06435}.

\bibitemdeclare{misc}{ng_deeplearningai_nodate}
\bibitem{ng_deeplearningai_nodate}
\bibinfo{author}{Andrew \surnamestart Ng\surnameend}:
  \emph{\bibinfo{title}{{DeepLearning}.{AI}}}.
\newblock \urlprefix\url{https://www.deeplearning.ai/}.

\bibitemdeclare{misc}{ning2024}
\bibitem{ning2024}
\bibinfo{author}{Lin \surnamestart Ning\surnameend}, \bibinfo{author}{Harsh
  \surnamestart Lara\surnameend}, \bibinfo{author}{Meiqi \surnamestart
  Guo\surnameend} \& \bibinfo{author}{Abhinav \surnamestart Rastogi\surnameend}
  (\bibinfo{year}{2024}): \emph{\bibinfo{title}{{MoDE: Effective Multi-task
  Parameter Efficient Fine-Tuning with a Mixture of Dyadic Experts}}}.
\newblock \urlprefix\url{https://arxiv.org/abs/2408.01505}.

\bibitemdeclare{misc}{ouyang2022}
\bibitem{ouyang2022}
\bibinfo{author}{Long \surnamestart Ouyang\surnameend}, \bibinfo{author}{Jeff
  \surnamestart Wu\surnameend}, \bibinfo{author}{Xu~\surnamestart
  Jiang\surnameend}, \bibinfo{author}{Diogo \surnamestart Almeida\surnameend},
  \bibinfo{author}{Carroll~L \surnamestart Wainwright\surnameend},
  \bibinfo{author}{Pamela \surnamestart Mishkin\surnameend},
  \bibinfo{author}{Chong \surnamestart Zhang\surnameend},
  \bibinfo{author}{Sandhini \surnamestart Agarwal\surnameend},
  \bibinfo{author}{Katarina \surnamestart Slama\surnameend},
  \bibinfo{author}{Alex \surnamestart Ray\surnameend}, \bibinfo{author}{John
  \surnamestart Schulman\surnameend}, \bibinfo{author}{Jacob \surnamestart
  Hilton\surnameend}, \bibinfo{author}{Fraser \surnamestart Kelton\surnameend},
  \bibinfo{author}{Luke \surnamestart Miller\surnameend},
  \bibinfo{author}{Maddie \surnamestart Simens\surnameend},
  \bibinfo{author}{Amanda \surnamestart Askell\surnameend},
  \bibinfo{author}{Peter \surnamestart Welinder\surnameend},
  \bibinfo{author}{Paul \surnamestart Christiano\surnameend},
  \bibinfo{author}{Jan \surnamestart Leike\surnameend} \& \bibinfo{author}{Ryan
  \surnamestart Lowe\surnameend} (\bibinfo{year}{2022}):
  \emph{\bibinfo{title}{{Training language models to follow instructions with
  human feedback}}}.
\newblock \urlprefix\url{https://arxiv.org/abs/2203.02155}.

\bibitemdeclare{article}{Panoutsopoulos2024}
\bibitem{Panoutsopoulos2024}
\bibinfo{author}{Hercules \surnamestart Panoutsopoulos\surnameend},
  \bibinfo{author}{Borja \surnamestart Espejo-Garcia\surnameend},
  \bibinfo{author}{Stephan \surnamestart Raaijmakers\surnameend},
  \bibinfo{author}{Xu~\surnamestart Wang\surnameend}, \bibinfo{author}{Spyros
  \surnamestart Fountas\surnameend} \& \bibinfo{author}{Christopher
  \surnamestart Brewster\surnameend} (\bibinfo{year}{2024}):
  \emph{\bibinfo{title}{{Investigating the effect of different fine-tuning
  configuration scenarios on agricultural term extraction using BERT}}}.
\newblock {\sl \bibinfo{journal}{Computers and Electronics in Agriculture}}
  \bibinfo{volume}{225}, p. \bibinfo{pages}{109268},
  \doi{https://doi.org/10.1016/j.compag.2024.109268}.
\newblock
  \urlprefix\url{https://www.sciencedirect.com/science/article/pii/S0168169924006598}.

\bibitemdeclare{misc}{parthasarathy2024}
\bibitem{parthasarathy2024}
\bibinfo{author}{Venkatesh~Balavadhani \surnamestart Parthasarathy\surnameend},
  \bibinfo{author}{Ahtsham \surnamestart Zafar\surnameend},
  \bibinfo{author}{Aafaq \surnamestart Khan\surnameend} \&
  \bibinfo{author}{Arsalan \surnamestart Shahid\surnameend}
  (\bibinfo{year}{2024}): \emph{\bibinfo{title}{{The Ultimate Guide to
  Fine-Tuning LLMs from Basics to Breakthroughs: An Exhaustive Review of
  Technologies, Research, Best Practices, Applied Research Challenges and
  Opportunities}}}.
\newblock \urlprefix\url{https://arxiv.org/abs/2408.13296}.

\bibitemdeclare{inproceedings}{PfeifferJonas2021ANtc}
\bibitem{PfeifferJonas2021ANtc}
\bibinfo{author}{Jonas \surnamestart Pfeiffer\surnameend},
  \bibinfo{author}{Aishwarya \surnamestart Kamath\surnameend},
  \bibinfo{author}{Andreas \surnamestart R{\"{u}}ckl{\'{e}}\surnameend},
  \bibinfo{author}{Kyunghyun \surnamestart Cho\surnameend} \&
  \bibinfo{author}{Iryna \surnamestart Gurevych\surnameend}
  (\bibinfo{year}{2021}): \emph{\bibinfo{title}{{AdapterFusion: Non-destructive
  task composition for transfer learning}}}.
\newblock In: {\sl \bibinfo{booktitle}{EACL 2021 - 16th Conference of the
  European Chapter of the Association for Computational Linguistics,
  Proceedings of the Conference}}, pp. \bibinfo{pages}{487--503}.

\bibitemdeclare{misc}{raffel_exploring_2023}
\bibitem{raffel_exploring_2023}
\bibinfo{author}{Colin \surnamestart Raffel\surnameend}, \bibinfo{author}{Noam
  \surnamestart Shazeer\surnameend}, \bibinfo{author}{Adam \surnamestart
  Roberts\surnameend}, \bibinfo{author}{Katherine \surnamestart
  Lee\surnameend}, \bibinfo{author}{Sharan \surnamestart Narang\surnameend},
  \bibinfo{author}{Michael \surnamestart Matena\surnameend},
  \bibinfo{author}{Yanqi \surnamestart Zhou\surnameend}, \bibinfo{author}{Wei
  \surnamestart Li\surnameend} \& \bibinfo{author}{Peter~J \surnamestart
  Liu\surnameend} (\bibinfo{year}{2023}): \emph{\bibinfo{title}{{Exploring the
  Limits of Transfer Learning with a Unified Text-to-Text Transformer}}},
  \doi{10.48550/arXiv.1910.10683}.
\newblock \urlprefix\url{http://arxiv.org/abs/1910.10683}.

\bibitemdeclare{misc}{rolnick_experience_2019}
\bibitem{rolnick_experience_2019}
\bibinfo{author}{David \surnamestart Rolnick\surnameend}, \bibinfo{author}{Arun
  \surnamestart Ahuja\surnameend}, \bibinfo{author}{Jonathan \surnamestart
  Schwarz\surnameend}, \bibinfo{author}{Timothy~P \surnamestart
  Lillicrap\surnameend} \& \bibinfo{author}{Greg \surnamestart
  Wayne\surnameend} (\bibinfo{year}{2019}): \emph{\bibinfo{title}{{Experience
  Replay for Continual Learning}}}, \doi{10.48550/arXiv.1811.11682}.
\newblock \urlprefix\url{http://arxiv.org/abs/1811.11682}.

\bibitemdeclare{article}{roodschild_new_2020}
\bibitem{roodschild_new_2020}
\bibinfo{author}{Mat{\'{i}}as \surnamestart Roodschild\surnameend},
  \bibinfo{author}{Jorge \surnamestart {Gotay Sardi{\~{n}}as}\surnameend} \&
  \bibinfo{author}{Adri{\'{a}}n \surnamestart Will\surnameend}
  (\bibinfo{year}{2020}): \emph{\bibinfo{title}{{A new approach for the
  vanishing gradient problem on sigmoid activation}}}.
\newblock {\sl \bibinfo{journal}{Progress in Artificial Intelligence}}
  \bibinfo{volume}{9}(\bibinfo{number}{4}), pp. \bibinfo{pages}{351--360},
  \doi{10.1007/s13748-020-00218-y}.
\newblock \urlprefix\url{https://doi.org/10.1007/s13748-020-00218-y}.

\bibitemdeclare{misc}{rusu_progressive_2022}
\bibitem{rusu_progressive_2022}
\bibinfo{author}{Andrei~A \surnamestart Rusu\surnameend},
  \bibinfo{author}{Neil~C \surnamestart Rabinowitz\surnameend},
  \bibinfo{author}{Guillaume \surnamestart Desjardins\surnameend},
  \bibinfo{author}{Hubert \surnamestart Soyer\surnameend},
  \bibinfo{author}{James \surnamestart Kirkpatrick\surnameend},
  \bibinfo{author}{Koray \surnamestart Kavukcuoglu\surnameend},
  \bibinfo{author}{Razvan \surnamestart Pascanu\surnameend} \&
  \bibinfo{author}{Raia \surnamestart Hadsell\surnameend}
  (\bibinfo{year}{2022}): \emph{\bibinfo{title}{{Progressive Neural
  Networks}}}, \doi{10.48550/arXiv.1606.04671}.
\newblock \urlprefix\url{http://arxiv.org/abs/1606.04671}.

\bibitemdeclare{inproceedings}{schoch_nl2ibe_2024}
\bibitem{schoch_nl2ibe_2024}
\bibinfo{author}{Nicolai \surnamestart Schoch\surnameend} \&
  \bibinfo{author}{Mario \surnamestart Hoernicke\surnameend}
  (\bibinfo{year}{2024}): \emph{\bibinfo{title}{{NL2IBE} –
  {Ontology}-controlled {Transformation} of {Natural} {Language} into
  {Formalized} {Engineering} {Artefacts}}}.
\newblock In: {\sl \bibinfo{booktitle}{2024 {IEEE} {Conference} on {Artificial}
  {Intelligence} ({CAI})}}, \bibinfo{publisher}{IEEE},
  \bibinfo{address}{Singapore, Singapore}, pp. \bibinfo{pages}{997--1004},
  \doi{10.1109/CAI59869.2024.00182}.
\newblock \urlprefix\url{https://ieeexplore.ieee.org/document/10605389/}.

\bibitemdeclare{article}{shrestha_post-translational_2024}
\bibitem{shrestha_post-translational_2024}
\bibinfo{author}{Palistha \surnamestart Shrestha\surnameend},
  \bibinfo{author}{Jeevan \surnamestart Kandel\surnameend},
  \bibinfo{author}{Hilal \surnamestart Tayara\surnameend} \&
  \bibinfo{author}{Kil~To \surnamestart Chong\surnameend}
  (\bibinfo{year}{2024}): \emph{\bibinfo{title}{{Post-translational
  modification prediction via prompt-based fine-tuning of a GPT-2 model}}}.
\newblock {\sl \bibinfo{journal}{Nature Communications}}
  \bibinfo{volume}{15}(\bibinfo{number}{1}), p. \bibinfo{pages}{6699},
  \doi{10.1038/s41467-024-51071-9}.
\newblock
  \urlprefix\url{https://www.proquest.com/docview/3089699098/abstract/562FFB7E6B4D4F0CPQ/1}.

\bibitemdeclare{misc}{Shumailov_Shumaylov_Zhao_Gal_Papernot_Anderson_2024}
\bibitem{Shumailov_Shumaylov_Zhao_Gal_Papernot_Anderson_2024}
\bibinfo{author}{Ilia \surnamestart Shumailov\surnameend},
  \bibinfo{author}{Zakhar \surnamestart Shumaylov\surnameend},
  \bibinfo{author}{Yiren \surnamestart Zhao\surnameend}, \bibinfo{author}{Yarin
  \surnamestart Gal\surnameend}, \bibinfo{author}{Nicolas \surnamestart
  Papernot\surnameend} \& \bibinfo{author}{Ross \surnamestart
  Anderson\surnameend} (\bibinfo{year}{2024}): \emph{\bibinfo{title}{The Curse
  of Recursion: Training on Generated Data Makes Models Forget}}.
\newblock \urlprefix\url{http://arxiv.org/abs/2305.17493}.
\newblock \bibinfo{note}{ArXiv:2305.17493 [cs]}.

\bibitemdeclare{book}{Smith2024Customer}
\bibitem{Smith2024Customer}
\bibinfo{author}{Ross \surnamestart Smith\surnameend},
  \bibinfo{author}{Mayte~Cubino \surnamestart Gonzalez\surnameend} \&
  \bibinfo{author}{Emily \surnamestart McKeon\surnameend}
  (\bibinfo{year}{2024}): \emph{\bibinfo{title}{{The AI Revolution in Customer
  Service and Support: A Practical Guide to Impactful Deployment of AI to Best
  Serve Your Customers}}}, \bibinfo{edition}{[first edi} edition.
\newblock \bibinfo{publisher}{Pearson}, \bibinfo{address}{Hoboken, New Jersey}.

\bibitemdeclare{article}{strachan_testing_2024}
\bibitem{strachan_testing_2024}
\bibinfo{author}{James W.~A. \surnamestart Strachan\surnameend},
  \bibinfo{author}{Dalila \surnamestart Albergo\surnameend},
  \bibinfo{author}{Giulia \surnamestart Borghini\surnameend},
  \bibinfo{author}{Oriana \surnamestart Pansardi\surnameend},
  \bibinfo{author}{Eugenio \surnamestart Scaliti\surnameend},
  \bibinfo{author}{Saurabh \surnamestart Gupta\surnameend},
  \bibinfo{author}{Krati \surnamestart Saxena\surnameend},
  \bibinfo{author}{Alessandro \surnamestart Rufo\surnameend},
  \bibinfo{author}{Stefano \surnamestart Panzeri\surnameend},
  \bibinfo{author}{Guido \surnamestart Manzi\surnameend},
  \bibinfo{author}{Michael S.~A. \surnamestart Graziano\surnameend} \&
  \bibinfo{author}{Cristina \surnamestart Becchio\surnameend}
  (\bibinfo{year}{2024}): \emph{\bibinfo{title}{Testing theory of mind in large
  language models and humans}}.
\newblock {\sl \bibinfo{journal}{Nat Hum Behav}}, pp. \bibinfo{pages}{1--11},
  \doi{10.1038/s41562-024-01882-z}.
\newblock \urlprefix\url{https://www.nature.com/articles/s41562-024-01882-z}.

\bibitemdeclare{misc}{tjondronegoro_strategic_2024}
\bibitem{tjondronegoro_strategic_2024}
\bibinfo{author}{Dian~W \surnamestart Tjondronegoro\surnameend}
  (\bibinfo{year}{2024}): \emph{\bibinfo{title}{Strategic {AI} {Governance}:
  {Insights} from {Leading} {Nations}}},
  \doi{https://doi.org/10.48550/arXiv.2410.01819}.
\newblock \urlprefix\url{https://doi.org/10.48550/arXiv.2410.01819}.

\bibitemdeclare{misc}{touvron2023llamaopenefficientfoundation}
\bibitem{touvron2023llamaopenefficientfoundation}
\bibinfo{author}{Hugo \surnamestart Touvron\surnameend},
  \bibinfo{author}{Thibaut \surnamestart Lavril\surnameend},
  \bibinfo{author}{Gautier \surnamestart Izacard\surnameend},
  \bibinfo{author}{Xavier \surnamestart Martinet\surnameend},
  \bibinfo{author}{Marie-Anne \surnamestart Lachaux\surnameend},
  \bibinfo{author}{Timoth{\'{e}}e \surnamestart Lacroix\surnameend},
  \bibinfo{author}{Baptiste \surnamestart Rozi{\`{e}}re\surnameend},
  \bibinfo{author}{Naman \surnamestart Goyal\surnameend}, \bibinfo{author}{Eric
  \surnamestart Hambro\surnameend}, \bibinfo{author}{Faisal \surnamestart
  Azhar\surnameend}, \bibinfo{author}{Aurelien \surnamestart
  Rodriguez\surnameend}, \bibinfo{author}{Armand \surnamestart
  Joulin\surnameend}, \bibinfo{author}{Edouard \surnamestart Grave\surnameend}
  \& \bibinfo{author}{Guillaume \surnamestart Lample\surnameend}
  (\bibinfo{year}{2023}): \emph{\bibinfo{title}{{LLaMA: Open and Efficient
  Foundation Language Models}}}.
\newblock \urlprefix\url{https://arxiv.org/abs/2302.13971}.

\bibitemdeclare{misc}{vaswani_attention_2023}
\bibitem{vaswani_attention_2023}
\bibinfo{author}{Ashish \surnamestart Vaswani\surnameend},
  \bibinfo{author}{Noam \surnamestart Shazeer\surnameend},
  \bibinfo{author}{Niki \surnamestart Parmar\surnameend},
  \bibinfo{author}{Jakob \surnamestart Uszkoreit\surnameend},
  \bibinfo{author}{Llion \surnamestart Jones\surnameend},
  \bibinfo{author}{Aidan~N. \surnamestart Gomez\surnameend},
  \bibinfo{author}{Lukasz \surnamestart Kaiser\surnameend} \&
  \bibinfo{author}{Illia \surnamestart Polosukhin\surnameend}
  (\bibinfo{year}{2023}): \emph{\bibinfo{title}{Attention {Is} {All} {You}
  {Need}}}, \doi{10.48550/arXiv.1706.03762}.
\newblock \urlprefix\url{http://arxiv.org/abs/1706.03762}.
\newblock \bibinfo{note}{ArXiv:1706.03762 [cs]}.

\bibitemdeclare{misc}{wang_what_2022}
\bibitem{wang_what_2022}
\bibinfo{author}{Thomas \surnamestart Wang\surnameend}, \bibinfo{author}{Adam
  \surnamestart Roberts\surnameend}, \bibinfo{author}{Daniel \surnamestart
  Hesslow\surnameend}, \bibinfo{author}{Teven~Le \surnamestart
  Scao\surnameend}, \bibinfo{author}{Hyung~Won \surnamestart Chung\surnameend},
  \bibinfo{author}{Iz~\surnamestart Beltagy\surnameend},
  \bibinfo{author}{Julien \surnamestart Launay\surnameend} \&
  \bibinfo{author}{Colin \surnamestart Raffel\surnameend}
  (\bibinfo{year}{2022}): \emph{\bibinfo{title}{What {Language} {Model}
  {Architecture} and {Pretraining} {Objective} {Work} {Best} for {Zero}-{Shot}
  {Generalization}?}}, \doi{10.48550/arXiv.2204.05832}.
\newblock \urlprefix\url{http://arxiv.org/abs/2204.05832}.
\newblock \bibinfo{note}{ArXiv:2204.05832 [cs, stat]}.

\bibitemdeclare{misc}{wei2022zeroshot}
\bibitem{wei2022zeroshot}
\bibinfo{author}{Jason \surnamestart Wei\surnameend}, \bibinfo{author}{Maarten
  \surnamestart Bosma\surnameend}, \bibinfo{author}{Vincent~Y \surnamestart
  Zhao\surnameend}, \bibinfo{author}{Kelvin \surnamestart Guu\surnameend},
  \bibinfo{author}{Adams~Wei \surnamestart Yu\surnameend},
  \bibinfo{author}{Brian \surnamestart Lester\surnameend}, \bibinfo{author}{Nan
  \surnamestart Du\surnameend}, \bibinfo{author}{Andrew~M \surnamestart
  Dai\surnameend} \& \bibinfo{author}{Quoc~V \surnamestart Le\surnameend}
  (\bibinfo{year}{2022}): \emph{\bibinfo{title}{{Finetuned Language Models Are
  Zero-Shot Learners}}}.
\newblock \urlprefix\url{https://arxiv.org/abs/2109.01652}.

\bibitemdeclare{misc}{wei2023chainofthoughtpromptingelicitsreasoning}
\bibitem{wei2023chainofthoughtpromptingelicitsreasoning}
\bibinfo{author}{Jason \surnamestart Wei\surnameend}, \bibinfo{author}{Xuezhi
  \surnamestart Wang\surnameend}, \bibinfo{author}{Dale \surnamestart
  Schuurmans\surnameend}, \bibinfo{author}{Maarten \surnamestart
  Bosma\surnameend}, \bibinfo{author}{Brian \surnamestart Ichter\surnameend},
  \bibinfo{author}{Fei \surnamestart Xia\surnameend},
  \bibinfo{author}{Ed~\surnamestart Chi\surnameend}, \bibinfo{author}{Quoc
  \surnamestart Le\surnameend} \& \bibinfo{author}{Denny \surnamestart
  Zhou\surnameend} (\bibinfo{year}{2023}):
  \emph{\bibinfo{title}{{Chain-of-Thought Prompting Elicits Reasoning in Large
  Language Models}}}.
\newblock \urlprefix\url{https://arxiv.org/abs/2201.11903}.

\bibitemdeclare{article}{Woo2024}
\bibitem{Woo2024}
\bibinfo{author}{David~James \surnamestart Woo\surnameend},
  \bibinfo{author}{Kai \surnamestart Guo\surnameend} \&
  \bibinfo{author}{Sdenka~Zobeida \surnamestart Salas-Pilco\surnameend}
  (\bibinfo{year}{2024}): \emph{\bibinfo{title}{{Writing creative stories with
  AI: learning designs for secondary school students}}}.
\newblock {\sl \bibinfo{journal}{Innovation in language learning and
  teaching}}, pp. \bibinfo{pages}{1--13}.

\bibitemdeclare{misc}{wu_bloomberggpt_2023}
\bibitem{wu_bloomberggpt_2023}
\bibinfo{author}{Shijie \surnamestart Wu\surnameend}, \bibinfo{author}{Ozan
  \surnamestart Irsoy\surnameend}, \bibinfo{author}{Steven \surnamestart
  Lu\surnameend}, \bibinfo{author}{Vadim \surnamestart Dabravolski\surnameend},
  \bibinfo{author}{Mark \surnamestart Dredze\surnameend},
  \bibinfo{author}{Sebastian \surnamestart Gehrmann\surnameend},
  \bibinfo{author}{Prabhanjan \surnamestart Kambadur\surnameend},
  \bibinfo{author}{David \surnamestart Rosenberg\surnameend} \&
  \bibinfo{author}{Gideon \surnamestart Mann\surnameend}
  (\bibinfo{year}{2023}): \emph{\bibinfo{title}{{BloombergGPT}: {A} {Large}
  {Language} {Model} for {Finance}}}.
\newblock \urlprefix\url{https://arxiv.org/abs/2303.17564v3}.

\bibitemdeclare{article}{xie2023}
\bibitem{xie2023}
\bibinfo{author}{Yueqi \surnamestart Xie\surnameend}, \bibinfo{author}{Jingwei
  \surnamestart Yi\surnameend}, \bibinfo{author}{Jiawei \surnamestart
  Shao\surnameend}, \bibinfo{author}{Justin \surnamestart Curl\surnameend},
  \bibinfo{author}{Lingjuan \surnamestart Lyu\surnameend},
  \bibinfo{author}{Qifeng \surnamestart Chen\surnameend}, \bibinfo{author}{Xing
  \surnamestart Xie\surnameend} \& \bibinfo{author}{Fangzhao \surnamestart
  Wu\surnameend} (\bibinfo{year}{2023}): \emph{\bibinfo{title}{{Defending
  ChatGPT against jailbreak attack via self-reminders}}}.
\newblock {\sl \bibinfo{journal}{Nature Machine Intelligence}}
  \bibinfo{volume}{5}(\bibinfo{number}{12}), pp. \bibinfo{pages}{1486--1496},
  \doi{10.1038/s42256-023-00765-8}.
\newblock \urlprefix\url{https://www.nature.com/articles/s42256-023-00765-8}.

\bibitemdeclare{misc}{xue2021mt5}
\bibitem{xue2021mt5}
\bibinfo{author}{Linting \surnamestart Xue\surnameend}, \bibinfo{author}{Noah
  \surnamestart Constant\surnameend}, \bibinfo{author}{Adam \surnamestart
  Roberts\surnameend}, \bibinfo{author}{Mihir \surnamestart Kale\surnameend},
  \bibinfo{author}{Rami \surnamestart Al-Rfou\surnameend},
  \bibinfo{author}{Aditya \surnamestart Siddhant\surnameend},
  \bibinfo{author}{Aditya \surnamestart Barua\surnameend} \&
  \bibinfo{author}{Colin \surnamestart Raffel\surnameend}
  (\bibinfo{year}{2021}): \emph{\bibinfo{title}{{mT5: A massively multilingual
  pre-trained text-to-text transformer}}}.
\newblock \urlprefix\url{https://arxiv.org/abs/2010.11934}.

\bibitemdeclare{inproceedings}{zafrir2019}
\bibitem{zafrir2019}
\bibinfo{author}{Ofir \surnamestart Zafrir\surnameend}, \bibinfo{author}{Guy
  \surnamestart Boudoukh\surnameend}, \bibinfo{author}{Peter \surnamestart
  Izsak\surnameend} \& \bibinfo{author}{Moshe \surnamestart
  Wasserblat\surnameend} (\bibinfo{year}{2019}): \emph{\bibinfo{title}{Q8BERT:
  Quantized 8Bit BERT}}.
\newblock In: {\sl \bibinfo{booktitle}{2019 Fifth Workshop on Energy Efficient
  Machine Learning and Cognitive Computing - NeurIPS Edition (EMC2-NIPS)}},
  \bibinfo{publisher}{IEEE}, pp. \bibinfo{pages}{36--39},
  \doi{10.1109/emc2-nips53020.2019.00016}.
\newblock \urlprefix\url{http://dx.doi.org/10.1109/EMC2-NIPS53020.2019.00016}.

\bibitemdeclare{inproceedings}{zhai_investigating_2024}
\bibitem{zhai_investigating_2024}
\bibinfo{author}{Yuexiang \surnamestart Zhai\surnameend},
  \bibinfo{author}{Shengbang \surnamestart Tong\surnameend},
  \bibinfo{author}{Xiao \surnamestart Li\surnameend},
  \bibinfo{author}{Mu~\surnamestart Cai\surnameend}, \bibinfo{author}{Qing
  \surnamestart Qu\surnameend}, \bibinfo{author}{Yong~Jae \surnamestart
  Lee\surnameend} \& \bibinfo{author}{Yi~\surnamestart Ma\surnameend}
  (\bibinfo{year}{2024}): \emph{\bibinfo{title}{{Investigating the catastrophic
  forgetting in multimodal large language model fine-tuning}}}.
\newblock In: {\sl \bibinfo{booktitle}{Conference on Parsimony and Learning}},
  \bibinfo{publisher}{PMLR}, pp. \bibinfo{pages}{202--227}.
\newblock \urlprefix\url{https://proceedings.mlr.press/v234/zhai24a.html}.

\bibitemdeclare{misc}{zhao_survey_2023}
\bibitem{zhao_survey_2023}
\bibinfo{author}{Wayne~Xin \surnamestart Zhao\surnameend}, \bibinfo{author}{Kun
  \surnamestart Zhou\surnameend}, \bibinfo{author}{Junyi \surnamestart
  Li\surnameend}, \bibinfo{author}{Tianyi \surnamestart Tang\surnameend},
  \bibinfo{author}{Xiaolei \surnamestart Wang\surnameend},
  \bibinfo{author}{Yupeng \surnamestart Hou\surnameend},
  \bibinfo{author}{Yingqian \surnamestart Min\surnameend},
  \bibinfo{author}{Beichen \surnamestart Zhang\surnameend},
  \bibinfo{author}{Junjie \surnamestart Zhang\surnameend},
  \bibinfo{author}{Zican \surnamestart Dong\surnameend}, \bibinfo{author}{Yifan
  \surnamestart Du\surnameend}, \bibinfo{author}{Chen \surnamestart
  Yang\surnameend}, \bibinfo{author}{Yushuo \surnamestart Chen\surnameend},
  \bibinfo{author}{Zhipeng \surnamestart Chen\surnameend},
  \bibinfo{author}{Jinhao \surnamestart Jiang\surnameend},
  \bibinfo{author}{Ruiyang \surnamestart Ren\surnameend},
  \bibinfo{author}{Yifan \surnamestart Li\surnameend}, \bibinfo{author}{Xinyu
  \surnamestart Tang\surnameend}, \bibinfo{author}{Zikang \surnamestart
  Liu\surnameend}, \bibinfo{author}{Peiyu \surnamestart Liu\surnameend},
  \bibinfo{author}{Jian-Yun \surnamestart Nie\surnameend} \&
  \bibinfo{author}{Ji-Rong \surnamestart Wen\surnameend}
  (\bibinfo{year}{2023}): \emph{\bibinfo{title}{A {Survey} of {Large}
  {Language} {Models}}}, \doi{10.48550/arXiv.2303.18223}.
\newblock \urlprefix\url{http://arxiv.org/abs/2303.18223}.
\newblock \bibinfo{note}{ArXiv:2303.18223 [cs]}.

\bibitemdeclare{article}{zirar2023}
\bibitem{zirar2023}
\bibinfo{author}{Araz \surnamestart Zirar\surnameend},
  \bibinfo{author}{Syed~Imran \surnamestart Ali\surnameend} \&
  \bibinfo{author}{Nazrul \surnamestart Islam\surnameend}
  (\bibinfo{year}{2023}): \emph{\bibinfo{title}{{Worker and workplace
  Artificial Intelligence (AI) coexistence: Emerging themes and research
  agenda}}}.
\newblock {\sl \bibinfo{journal}{Technovation}} \bibinfo{volume}{124}, p.
  \bibinfo{pages}{102747},
  \doi{https://doi.org/10.1016/j.technovation.2023.102747}.
\newblock
  \urlprefix\url{https://www.sciencedirect.com/science/article/pii/S0166497223000585}.

\end{thebibliography}

\end{document}